\documentclass[sigconf, screen=true, review=false, printacmref=false, printccs=false, printfolios=false]{acmart}

\setcopyright{none}

\newtheorem{Lemma}{Lemma}[section]  
\newtheorem{Definition}{Definition}
\newtheorem{Claim}{Claim}[section]  

\usepackage{booktabs} 
\usepackage{enumitem}

\usepackage{subfigure}
\usepackage{rotating}

\usepackage{multirow}

\usepackage{comment}

\usepackage{color}
\usepackage{xcolor}
\definecolor{mydarkblue}{RGB}{0, 20, 159} 
\definecolor{mydarkblue}{rgb}{0,0.08,0.45} 
\usepackage{hyperref}
\hypersetup{colorlinks=true,urlcolor=mydarkblue, 
citecolor=mydarkblue,  linkcolor=mydarkblue}

\DeclareSymbolFont{cmbrightop}{OT1}{cmbr}{m}{n}
\DeclareMathSymbol{\sfPsi}{\mathalpha}{cmbrightop}{9}

\let\hat\widehat

\usepackage{balance}

\usepackage{epsfig}

\usepackage{graphicx}

\usepackage{amssymb}
\usepackage{amsmath}
\usepackage{amsfonts}

\usepackage{tabularx}
\usepackage{booktabs}
\usepackage{relsize}

\usepackage{numprint} 

\usepackage{setspace}
\graphicspath{{./}{./graphics/}}
\newcolumntype{H}{>{\setbox0=\hbox\bgroup}c<{\egroup}@{}}

\newcommand{\etal}{\emph{et al.}}

\newcommand{\eg}{\emph{e.g.}}
\newcommand{\ie}{\emph{i.e.}}

\usepackage{algorithm}
\usepackage{algorithmicx}
\usepackage{algpseudocode}
\algblockdefx[parallel]{ParFor}{EndPar}[1][]{$\textbf{parallel for}$ #1 $\textbf{do}$}{$\textbf{end parallel}$}
\algrenewcommand{\alglinenumber}[1]{\fontsize{6.5}{7}\selectfont#1}
\algtext*{EndPar}

\algblockdefx[parallel]{parfor}{endpar}[1][]{$\textbf{parallel for}$ #1 $\textbf{do}$}{$\textbf{end parallel}$}
\algrenewcommand{\alglinenumber}[1]{\scriptsize#1:}

\algblockdefx[parallel]{parallelfor}{parallelend}
[1][]{\textbf{parallel for} #1}
{\textbf{end parallel}}

\usepackage{nicefrac}

\algnotext{EndIf}
\algnotext{EndProcedure}

\usepackage{comment}
\usepackage{tabularx}

\usepackage{subfigure}
\usepackage{enumitem}

\usepackage{xcolor}
\definecolor{thedarkblue}{RGB}{0,0,120} 
\usepackage{hyperref}

\definecolor{mydarkblue}{rgb}{0,0.08,0.45} 
\hypersetup{
colorlinks=true,
linkcolor=mydarkblue,
citecolor=mydarkblue,
filecolor=mydarkblue,
urlcolor=mydarkblue,
pdfview=FitH}

\usepackage{microtype}

\usepackage{balance}

\renewcommand{\subsubsection}[1]{\medskip\noindent\textbf{#1}:}
\providecommand{\mat}[1]{\boldsymbol{\mathrm{#1}}}
\renewcommand{\vec}[1]{\boldsymbol{\mathrm{#1}}}

\newcommand{\argmin}{\operatornamewithlimits{argmin}}

\DeclareMathOperator{\hugeE}{\mbox{\huge\raise-0.3ex\hbox{E}}}
\DeclareMathOperator{\p}{\mathbb{P}}
\DeclareMathOperator{\hugep}{\mbox{\huge\raise-0.3ex\hbox{$\p$}}}

\newcommand{\RR}{\mathbb{R}}

\providecommand{\eye}{\mat{I}}
\providecommand{\mA}{\ensuremath{\mat{A}}}

\providecommand{\mD}{\ensuremath{\mat{D}}}

\providecommand{\mH}{\ensuremath{\mat{H}}}

\providecommand{\mL}{\ensuremath{\mat{L}}}

\providecommand{\mP}{\ensuremath{\mat{P}}}

\providecommand{\mS}{\ensuremath{\mat{S}}}

\providecommand{\mU}{\ensuremath{\mat{U}}}
\providecommand{\mV}{\ensuremath{\mat{V}}}
\providecommand{\mW}{\ensuremath{\mat{W}}}
\providecommand{\mX}{\ensuremath{\mat{X}}}
\providecommand{\mY}{\ensuremath{\mat{Y}}}
\providecommand{\mZ}{\ensuremath{\mat{Z}}}

\providecommand{\ve}{\ensuremath{\vec{e}}}

\providecommand{\vp}{\ensuremath{\vec{p}}}

\providecommand{\vx}{\ensuremath{\vec{x}}}

\providecommand{\vz}{\ensuremath{\vec{z}}}

\begin{document}

\title{HONE: Higher-Order Network Embeddings}

\author{Ryan A. Rossi}
\affiliation{
\institution{Adobe Research}
}
\email{}
\author{Nesreen K. Ahmed}
\affiliation{
\institution{Intel Labs}
}
\author{Eunyee Koh}
\affiliation{
\institution{Adobe Research}
}

\author{Sungchul Kim}
\affiliation{
\institution{Adobe Research}
}
\email{}
\author{Anup Rao}
\affiliation{
\institution{Adobe Research}
}
\email{}
\author{Yasin Abbasi Yadkori}
\affiliation{
\institution{Adobe Research}
}
\email{}

\renewcommand{\shortauthors}{R.~A.~Rossi et al.}

\begin{abstract}
Learning a useful representation from graph data lies at the heart and success of many machine learning tasks such as entity resolution, link prediction, user modeling, anomaly detection, and many others.
Recent methods mainly focus on learning graph representations based on random walks.
These methods are unable to capture higher-order dependencies and connectivity patterns that are crucial to understanding and modeling complex networks.
In this work, we formulate \emph{higher-order network representation learning} and describe a general framework for learning 
\emph{Higher-Order Network Embeddings} (HONE) from graph data based on lower-order subgraph patterns called graphlets (network motifs, orbits).
The HONE framework is highly expressive and flexible with many interchangeable components.
The experimental results demonstrate the effectiveness of learning higher-order network representations.
In all cases, HONE outperforms recent embedding methods that are unable to capture higher-order structures.
In particular, HONE achieves a mean relative gain in AUC of $19\%$ 
(and up to $75\%$ gain) across all methods and over a wide variety of networks from different application domains.
\end{abstract}

\begin{abstract}
This paper describes a general framework for learning \emph{Higher-Order Network Embeddings} (HONE) from graph data based on network motifs. The HONE framework is highly expressive and flexible with many interchangeable components. The experimental results demonstrate the effectiveness of learning higher-order network representations. In all cases, HONE outperforms recent embedding methods that are unable to capture higher-order structures with a mean relative gain in AUC of $19\%$ (and up to $75\%$ gain) across a wide variety of networks and embedding methods.
\end{abstract}




\maketitle

\section{Introduction} \label{sec:intro}
Roles represent node (or edge~\cite{ahmed2017roles})
connectivity patterns such as hub/star-center nodes, star-edge nodes, near-cliques or bridge nodes connecting different regions of the graph.
Intuitively, two nodes belong to the same role if they are structurally similar (with respect to their general connectivity/subgraph patterns)~\cite{roles2015-tkde}.
Informally, roles are sets of nodes that are more \emph{structurally similar} to nodes inside the set than outside, whereas communities are sets of nodes with more connections inside the set than outside.
Roles are complimentary but fundamentally different to the notion of communities.
Communities capture cohesive/tightly-knit groups of nodes and nodes in the same community are close together (small graph distance)~\cite{fortunato2010community}, whereas roles capture nodes that are structurally similar with respect to their general connectivity and subgraph patterns and are independent of the distance/proximity to one another in the graph~\cite{roles2015-tkde}.
Hence, two nodes that share similar roles can be in different communities and even in two disconnected components of the graph.
The goal of role learning in graphs is to not only group structurally similar nodes into sets but also to embed them close together in some $D$-dimensional space~\cite{roles2015-tkde}.

Many network representation learning methods attempt to capture the notion of roles (structural similarity)~\cite{roles2015-tkde} using random walks that are fundamentally tied to node identity and not general structural/subgraph patterns (network motifs) of nodes.
As such, two nodes with similar embeddings are guaranteed to be near one another in the graph (a property of communities~\cite{fortunato2010community}) since they appear near one another in a random walk.\footnote{Nodes in different disconnected components will never appear in a random walk together and therefore will not be assigned similar embeddings despite the fact these nodes may play the same roles with respect to general structural patterns such as network motifs.}
However, such methods are insufficient for roles~\cite{roles2015-tkde} as they fail to capture the general higher-order connectivity patterns of a node.
Moreover, past approaches that leverage traditional random walks (using node ids as opposed to attributed random walks that use ``types''~\cite{ahmed17Gen-Deep-Graph-Learning}) capture communities in the graph as opposed to node roles which are independent of the distance/proximity of nodes and instead represent higher-order connectivity patterns such as nodes that represent hubs or near-cliques.
For instance, instead of representing hub nodes (\eg, large star-centers) in a similar fashion,
methods using explicit random-walks (proximity/distance-based) 
would represent a hub node (star center) and its neighbors (star-edge) similarly despite them having fundamentally different connectivity patterns.

In this work, we propose \emph{higher-order network representation learning} and describe a general framework called 
Higher-Order Network Embeddings (HONE) for learning such higher-order embeddings based on network motifs.
The term motif is used generally and may refer to graphlets or orbits (graphlet automorphisms)~\cite{prvzulj2007biological,pgd}.
The approach leverages all available motif counts (and more generally statistics) by deriving a weighted motif graph $\mW_t$ 
from each network motif $H_t \in \mathcal{H}$ and uses these as a basis to learn higher-order embeddings that capture the notion of structural similarity (roles)~\cite{roles2015-tkde}.
The HONE framework expresses a new class of embedding methods based on a set of motif-based matrices and their powers.
In this work, we investigate HONE variants based on the weighted motif graph, motif transition matrix, motif Laplacian matrix, as well as other motif-based matrices.
The experiments demonstrate the effectiveness of HONE as we achieve a mean relative gain in AUC of $19\%$ across a variety of different networks and embedding methods.

\medskip\noindent\textbf{Contributions}:
This work makes three important contributions.
First, we introduce the problem of higher-order network representation learning.
Second, we propose a general class of methods for learning higher-order network embeddings based on network motifs.
Third, we demonstrate the effectiveness of learning higher-order network representations.

\section{Higher-Order Network Embeddings} \label{sec:framework}
This section proposes a new class of embedding models called \emph{Higher-Order Network Embeddings} (HONE) and a general framework for deriving them.
The class of higher-order network embedding methods is defined as follows:
\begin{Definition}[Higher-Order Network Embeddings] \label{def:higher-order-network-embedding} 
Given a network (graph) $G=(V,E)$, a set of network motifs $\mathcal{H} = \{H_1, \ldots, H_T\}$, 
the goal of higher-order network embedding (HONE) is to learn a function 
$f : V \rightarrow \RR^{D}$ that maps nodes to $D$-dimensional embeddings 
using network motifs $\mathcal{H}$.
\end{Definition}\noindent
The particular family of higher-order node embeddings presented in this work is based on learning a function $f : V \rightarrow \RR^{D}$ that maps nodes to $D$-dimensional embeddings using (powers of) weighted motif graphs derived from a motif matrix function $\Psi$.
However, many other families of higher-order node embedding methods exist in the class of higher-order network embeddings 
(Definition~\ref{def:higher-order-network-embedding}).
Most importantly, since network motifs lie at the heart of \emph{higher-order network embeddings} (Definition~\ref{def:higher-order-network-embedding}), they are guaranteed to capture the notion of roles (based on general subgraph/connectivity patterns of nodes)~\cite{roles2015-tkde} as opposed to the complimentary but fundamentally different notion of communities (based on proximity/small graph distance, and cohesive/tightly-knit/dense groups of nodes)~\cite{fortunato2010community}.

\subsection{Network Motifs}\label{sec:network-motifs}
\begin{figure}[h!]
\centering
\includegraphics[width=0.55\linewidth]{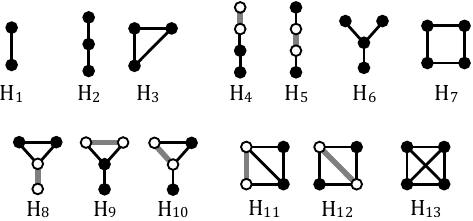}
\caption{All (2-4)-vertex connected edge orbits.
}
\label{fig:all-conn-edge-orbits-up-to-4-nodes}
\end{figure}

\noindent
The HONE framework can use graphlets or orbits.
Recall that the term motif is used generally in this work and may refer to graphlets or orbits (graphlet automorphisms)~\cite{prvzulj2007biological,pgd}.
\begin{Definition}[{\sc Graphlet}]\label{def:graphlet}
A graphlet $H_t = (V_k,E_k)$ is an induced subgraph consisting of a subset $V_k \subset V$ of $k$ vertices from $G = (V,E)$ together with all edges whose endpoints are both in this subset $E_k = \{ \forall e \in E \,|\, e = (u,v) \wedge u,v \in V_k \}$.
\end{Definition}\noindent
A $k$-graphlet is defined as an \emph{induced subgraph} with $k$-vertices.
Alternatively, the nodes of every graphlet can be partitioned into a set of automorphism groups called orbits~\cite{prvzulj2007biological}.
It is important to consider the \emph{position} of an edge in a graphlet, for instance, an edge in the 4-node path (Figure~\ref{fig:all-conn-edge-orbits-up-to-4-nodes}) has two different unique positions, namely, the edge in the center of the path, or an edge on the outside of the 4-node path. 
Each unique edge position in a graphlet is called an \emph{automorphism orbit}, or just orbit.
More formally,
\begin{Definition}[{\sc Orbit}]\label{def:orbit}
An automorphism of a $k$-vertex graphlet $H_t = (V_k,E_k)$ is defined as a permutation of the nodes in $H_t$ that preserves edges and non-edges. 
The automorphisms of $H_t$ form an automorphism group denoted as $Aut(H_t)$.
A set of nodes $V_k$ of graphlet $H_t$ define an \emph{orbit} iff 
(i) for any node $u \in V_k$ and any automorphism $\pi$ of $H_t$, $u \in V_k \Longleftrightarrow \pi(u) \in V_k$; and
(ii) if $v,u \in V_k$ then there exists an automorphism $\pi$ of $H_t$ and a $\gamma > 0$ such that $\pi^{\gamma}(u)=v$.
\end{Definition}\noindent
In this work, we use all (2-4)-vertex connected edge orbits and denote this set as $\mathcal{H}$.
For an example, Figure~\ref{fig:all-conn-edge-orbits-up-to-4-nodes} shows the connected edge orbits with up to 4-nodes.

\subsection{Weighted Motif Graphs}
\label{sec:weighted-motif-adj-matrix}
\begin{figure}[h!]
\centering
\begin{minipage}[b]{0.54\linewidth} 
\subfigure[Initial graph]{
\hspace{-2mm}
\includegraphics[width=1.00\linewidth]{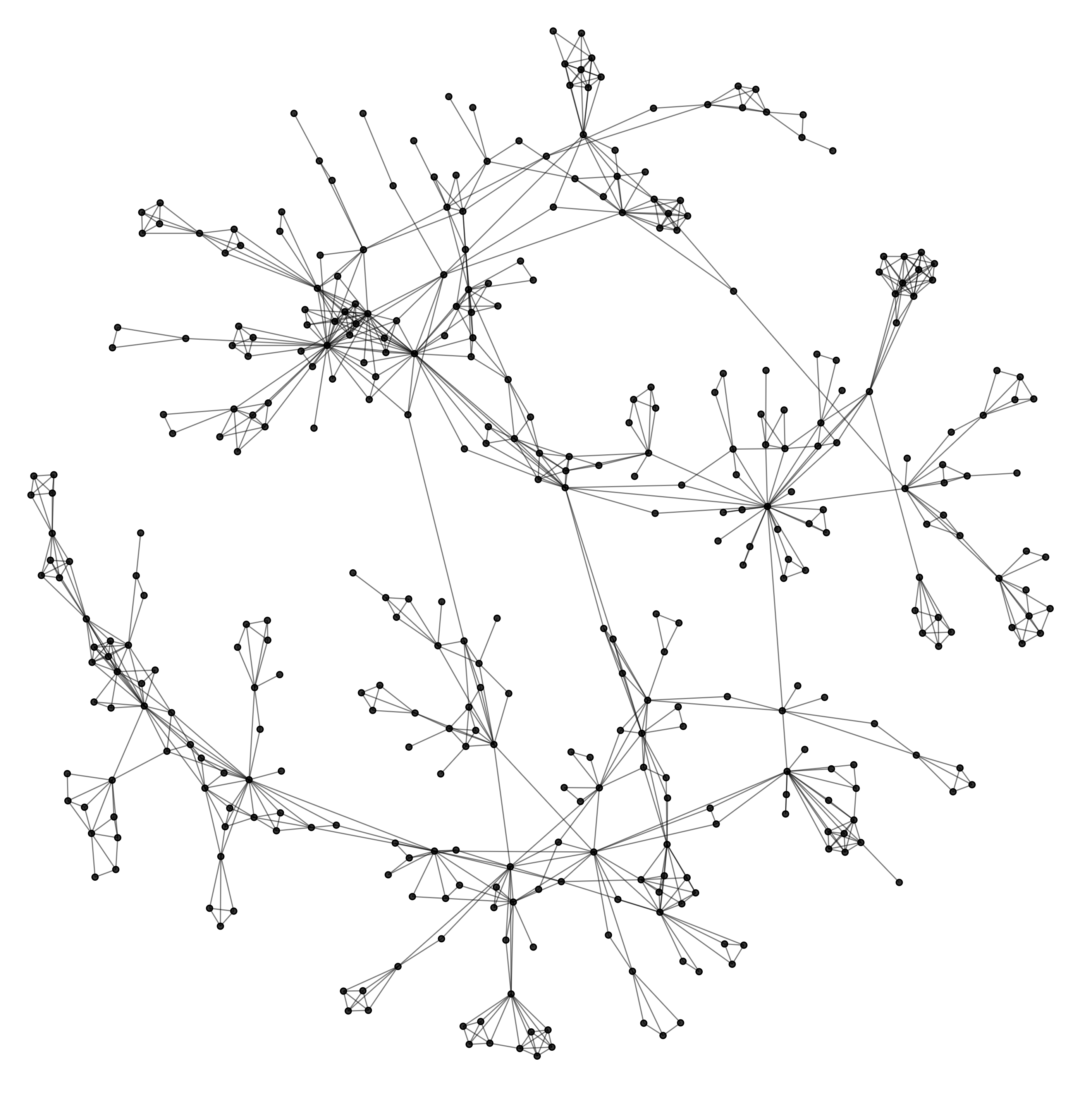} 
}
\vspace{2.5mm}
\end{minipage}
\begin{minipage}[b]{0.45\linewidth}
\hspace{4mm}
\subfigure[Weighted 4-clique graph]{\includegraphics[width=0.70\linewidth]{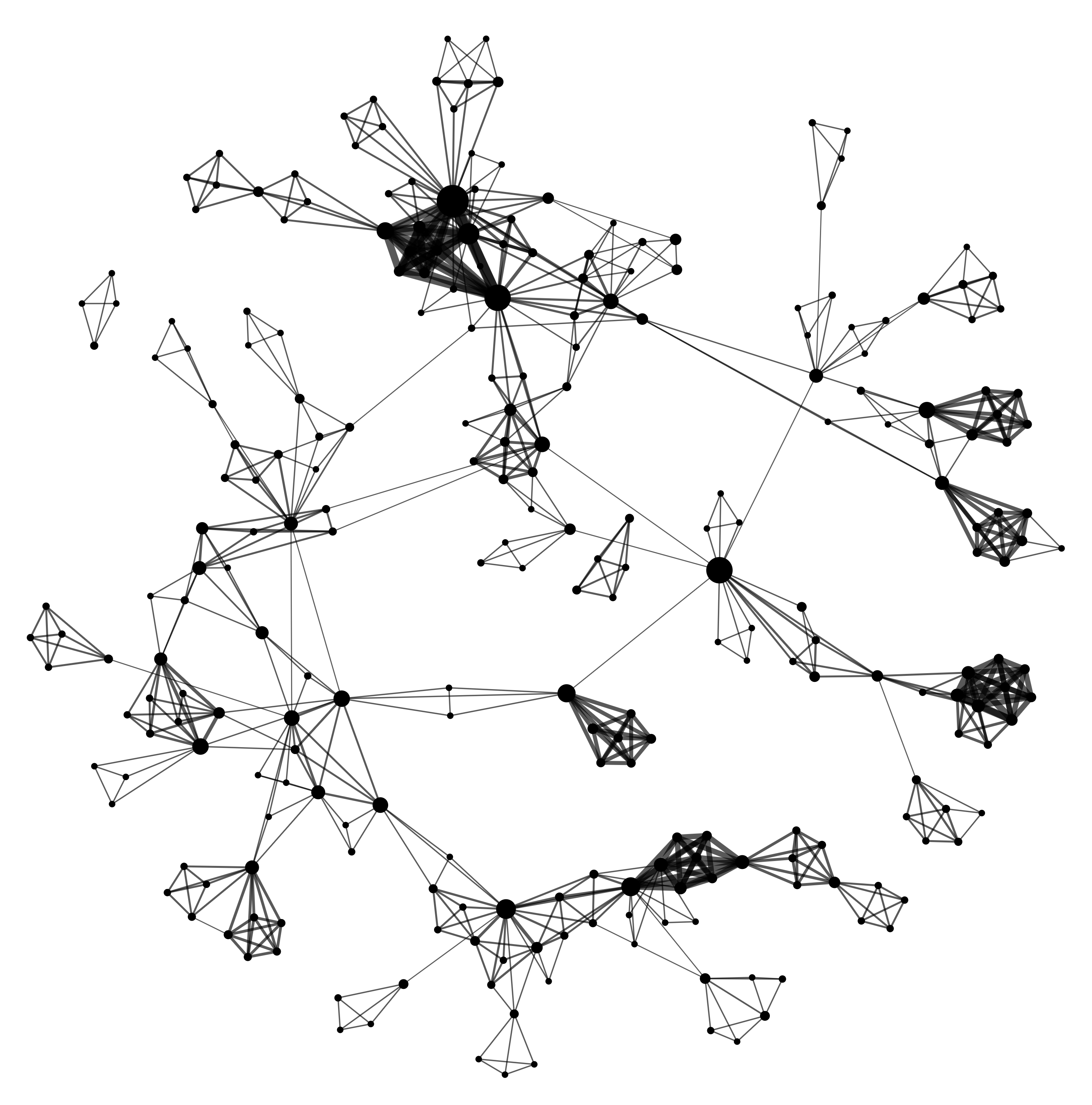}
}

\vspace{-3mm}
\hspace{4mm}
\subfigure[Weighted 4-path graph]{\includegraphics[width=0.70\linewidth]{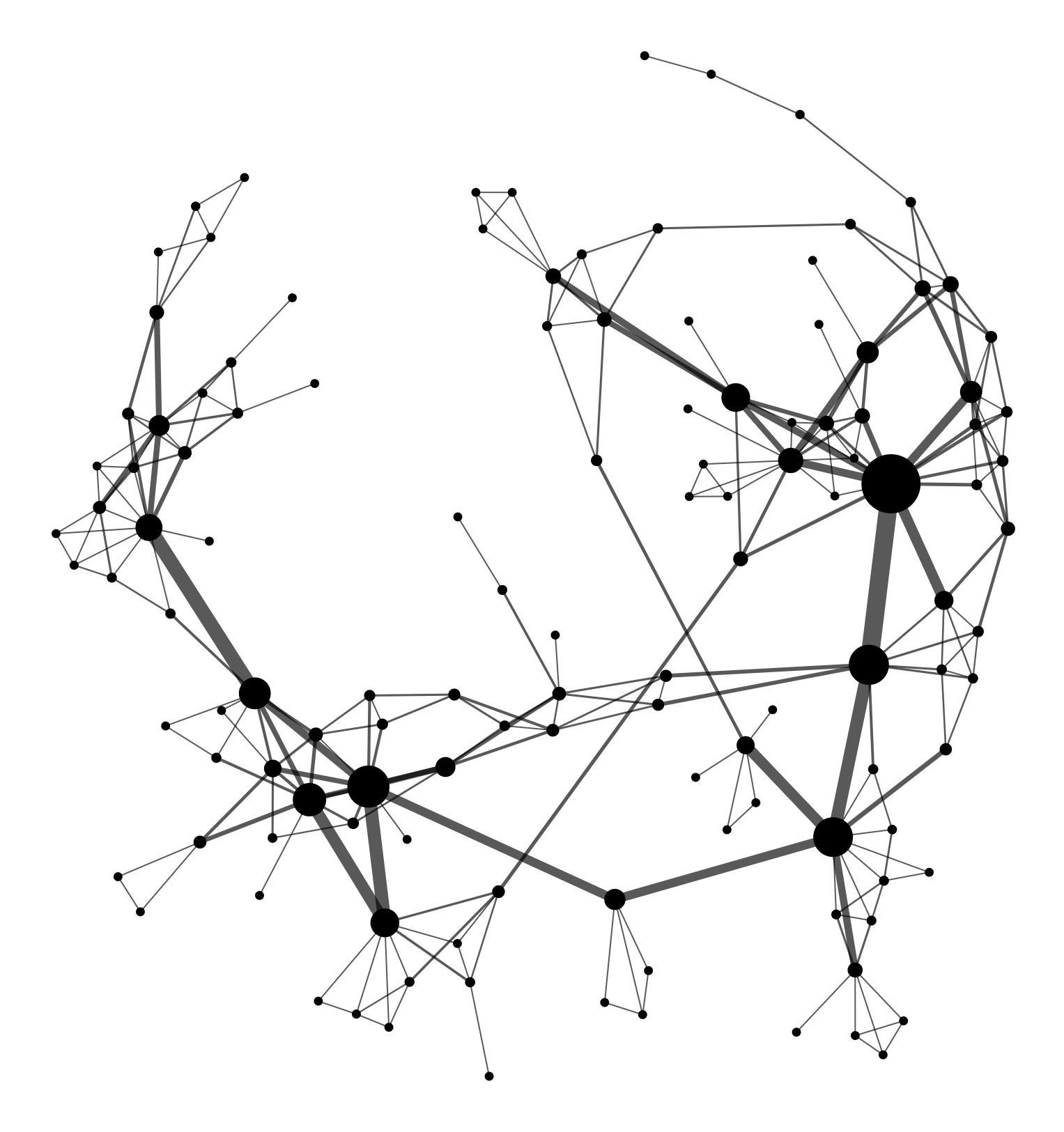}
}
\end{minipage}
\vspace{-2mm}
\caption{
Motif graphs differ in structure \textit{and} weight.
Size (weight) of nodes and edges in the 4-clique and 4-path motif graphs correspond to the frequency of 4-node cliques and 4-node paths, respectively.
}
\label{fig:motif-graph-comparison-ca-netscience}
\end{figure}

Given a network $G=(V,E)$ with $N=|V|$ nodes, $M=|E|$ edges, and a set $\mathcal{H} = \{H_1, \ldots, H_T\}$ of $T$ network motifs, we form the weighted motif adjacency matrices:
\begin{equation} \label{eq:motif-adj-matrices}
\mathcal{W} = \big\lbrace \mW_1, \mW_2, \ldots, \mW_T \big\rbrace
\end{equation} \noindent
where 
\begin{equation} \label{eq:motif-edge-weight}
(\mW_t)_{ij} = \# \text { occurences of motif } H_t \in \mathcal{H} \text{ that contain } (i, j) \in E \nonumber
\smallskip
\end{equation}\noindent
The weighted motif graphs differ from the original graph
in two important and fundamental ways.
First, the edges in each motif graph is likely to be weighted differently.
This is straightforward to see as each network motif can appear at a different frequency than another arbitrary motif for a given edge.
Intuitively, the edge motif weights when combined with the structure of the graph reveal important structural properties with respect to the weighted motif graph.
Second, the motif graphs are likely to be \emph{structurally} different (Figure~\ref{fig:motif-graph-comparison-ca-netscience}). 
For instance, if edge $(i,j) \in E$ exists in the original graph $G$, but $(\mW_t)_{ij}=0$ for some arbitrary motif $H_t$, then $(i,j) \not \in E_t$. 
Hence, the motif graphs encode relationships between nodes that have a sufficient number of motifs.
To generalize the above weighted motif graph formulation, we replace the edge constraint that says an edge exists between $i$ and $j$ if the number of instances of motif $H_t \in \mathcal{H}$ that contain nodes $i$ and $j$ is 1 or larger, by enforcing an edge constraint that requires each edge to have at least $\delta$ motifs.
In other words, different motif graphs can arise using the same motif $H_t$ by enforcing an edge constraint that requires each edge to have at least $\delta$ motifs.
This is an important property of the above formulation.

\subsection{Motif Matrix Functions} \label{sec:motif-based-matrix-functions}
To generalize HONE for any \emph{motif-based matrix formulation}, we define $\Psi$ as a function $\Psi : \RR^{N \times N} \! \rightarrow \RR^{N \times N}$ over a weighted motif adjacency matrix $\mW_t \in \mathcal{W}$.
Using $\Psi$ we derive
\begin{equation} \label{eq:motif-based-matrix}
\mS_t \,=\,  \Psi(\mW_t),\quad \text{for } t=1,2,\ldots,T
\end{equation}\noindent
The term \emph{motif-based matrix} refers to any motif matrix $\mS$ derived from $\Psi(\mW)$.\footnote{For convenience, $\mW$ denotes a weighted adjacency matrix for an arbitrary motif.}
We summarize the motif matrix functions $\Psi$ investigated below.
{
\begin{itemize}[leftmargin=1.1em]
\itemsep6pt

\item \textbf{Weighted Motif Graph}: 
Given a network $G$ and a network motif $H_t \in \mathcal{H}$,
form the weighted motif adjacency matrix $\mW_t$ whose entries $(i, j)$ are the co-occurrence counts of nodes $i$ and $j$ in the motif 
$H_t\!: (\mW_t)_{ij}$ = number of instances of $H_t$ that contain nodes $i$ and $j$.
In the case of using HONE directly with a weighted motif adjacency matrix $\mW$, then 
\begin{equation} \label{eq:motif-weighted-graph}
\Psi : \mW \rightarrow \eye \mW
\end{equation}
The number of paths weighted by motif counts from node $i$ to node $j$ in $k$-steps is given by
\begin{equation}\label{eq:motif-weighted-adjacency-matrix-W-node-ij-ksteps}
(\mW^k)_{ij} \,\, = \; \big( \,\underbrace{\mW \, \cdots \, \mW}_{k}\,\big)_{ij}
\end{equation}\noindent

\item \textbf{Motif Transition Matrix}: 
The random walk on a graph $\mW$ weighted by motif counts has transition probabilities 
\begin{equation}\label{eq:motif-transition-prob-single-edge-ij}
P_{ij} = \frac{W_{ij}}{w_i}
\end{equation}\noindent
where $w_i = \sum_{j} W_{ij}$ is the motif degree of node $i$.
The random walk motif transition matrix $\mP$ for an arbitrary weighted motif graph $\mW$ is defined as:
\begin{equation} \label{eq:motif-transition-prob-matrix} 
\mP =  \mD^{-1}\mW
\end{equation}\noindent
where $\mD = \mathtt{diag}(\mW \ve) = \mathtt{diag}(w_1, w_2, \ldots, w_N)$  is a $N \times N$ diagonal matrix with the motif degree $w_i = \sum_{j} W_{ij}$ of each node on the diagonal called the \emph{diagonal motif degree matrix} and $\ve = \big[ \, 1 \; 1\; \cdots \; 1 \, \big]^T$ is the vector of all ones. 
$\mP$ is a row-stochastic matrix with $\sum_{j} P_{ij} = \vp_{i}^T \ve = 1$ where $\vp_{i} \in \RR^{N}$ is a column vector corresponding to the $i$-th row of $\mP$.
For directed graphs, the motif out-degree is used. 
However, one can also leverage the motif in-degree or total motif degree (among other quantities).
The motif transition matrix $\mP$ represents the transition probabilities of a non-uniform random walk on the graph that selects subsequent nodes with probability proportional to the connecting edge's motif count.
Therefore, the probability of transitioning from node $i$ to node $j$ depends on the motif degree of $j$ relative to the total sum of motif degrees of all neighbors of $i$.
The probability of transitioning from node $i$ to $j$ in $k$-steps is given by 
\begin{equation}\label{eq:motif-transition-matrix-P-node-ij-ksteps}
(\mP^k)_{ij} 
\,\, = \; \big( \,\underbrace{\mP \,\, \cdots \,\, \mP}_{k}\,\big)_{ij}
\end{equation}

\item \textbf{Motif Laplacian}:
The motif Laplacian for a weighted \emph{motif} graph $\mW$ is defined as:
\begin{equation}\label{eq:simple-motif-laplacian}
\mL \, = \, \mD - \mW
\end{equation}\noindent
where $\mD = \mathtt{diag}(\mW\ve)$ is the diagonal motif degree matrix defined as $D_{ii} = \sum_{j} W_{ij}$.
For directed graphs, we can use either in-motif degree or out-motif degree.

\item \textbf{Normalized \emph{Motif} Laplacian}: 
Given a graph $\mW$ weighted by the counts of an arbitrary network motif $H_t \in \mathcal{H}$, the \emph{normalized motif Laplacian} is defined as
\begin{equation}\label{eq:norm-motif-laplacian}
\hat{\mL} \,\, = \,\, \eye \, - \, \mD^{-1/2} \mW \mD^{-1/2}
\end{equation}\noindent
where $\eye$ is the identity matrix and $\mD = \mathtt{diag}(\mW\ve)$ is the $N \times N$ diagonal matrix of motif degrees.
In other words,
\begin{equation} \label{eq:motif-norm-lap-full}
\hat{L}_{ij} = 
\begin{cases}
1 - \frac{W_{ij}}{w_j} & \text{if } i=j \text{ and } w_j \not= 0 \\
- \frac{W_{ij}}{ \sqrt{w_i w_j} } & \text{if } i \text{ and } j \text{ are adjacent} \\
0 & \text{otherwise} \\
\end{cases}
\end{equation}\noindent
where $w_i = \sum_{j} W_{ij}$ is the motif degree of node $i$.

\item \textbf{Random Walk Normalized Motif Laplacian}:
Formally, the \emph{random walk normalized motif Laplacian} is 
\begin{equation} \label{eq:norm-motif-laplacian-rw}
\hat{\mL}_{\rm rw} \; = \; \eye \, - \, \mD^{-1} \mW
\end{equation}
\noindent
where $\eye$ is the identity matrix, 
$\mD$ is the motif degree diagonal matrix with $D_{ii} = w_i, \forall i=1,\ldots,N$,
and $\mW$ is the weighted motif adjacency matrix for an arbitrary motif $H_t \in \mathcal{H}$.
Observe that $\hat{\mL}_{\rm rw} = \eye - \mP$ where $\mP = \mD^{-1}\mW$ is the motif transition matrix of a random walker on the weighted motif graph.

\medskip
\end{itemize}\noindent
}\noindent
Notice that all variants are easily formulated as functions $\Psi$ in terms of an arbitrary motif weighted graph $\mW$.

\subsection{Local K-Step Motif-based Embeddings} \label{sec:framework-local-Y}
We describe the \emph{local} higher-order node embeddings learned for each network motif $H_t \in \mathcal{H}$ and $k$-step where $k \in \{1,\ldots, K\}$.
The term local refers to the fact that node embeddings are learned for each individual motif and k-step independently.
We define $k$-step motif-based matrices for all $T$ motifs and $K$ steps as follows:
\begin{equation} \label{eq:motif-k-step-matrices}
\mS_t^{(k)} = \Psi(\mW^{k}_{t}),\;\, \text{ for } k=1,\ldots,K   \,\text{  and  }\,   t=1,\ldots,T 
\end{equation}
\noindent
where
\begin{equation} \label{eq:motif-k-step-matrices-expanded}
\Psi(\mW_t^k) \; = \; \,\Psi(\underbrace{\mW_t \,\, \cdots \,\, \mW_t}_{k})\;\,\;\;\;\\ 
\end{equation}\noindent
Note for the proposed motif Laplacian HONE variants $\mS^{(k)} \, = \;\, \Psi \big(\! \mW^{k} \big)$ ensures $\mS^{(k)}$ is a valid motif Laplacian matrix.
However, the motif transition probability matrix $\mP$ remains a valid transition matrix when taking powers of it and therefore we can simply use $\mS^{(k)} \, = \;\, \Psi \big( \mW \big)^k$ where \textcolor{black}{$\Psi : \mW \rightarrow \mD^{-1}\mW$}.
Depending on the motif-based matrix formulation $\Psi$ (Section~\ref{sec:motif-based-matrix-functions}), we renormalize each $k$-step motif matrix appropriately.
Alternatively, we can define
$\mS^{(k)} \, = \;\, \Psi \big( \mW \big)^k$ where we first use the motif matrix function $\Psi$ and then derive powers of the resulting motif-based matrix $\Psi(\mW)$.
Hence, 
\begin{equation} \label{eq:motif-k-step-matrices-expanded-prior}
\Psi(\mW_t)^{k} \; = \; \,\underbrace{\Psi(\mW_t) \,\, \cdots \,\, \Psi(\mW_t)}_{k}\;,\,\;\;\;\\ 
\end{equation}\noindent

\noindent
These k-step motif-based matrices can densify quickly and therefore the space required to store the k-step motif-based matrices can grow fast as $K$ increases.
For large graphs, it is often impractical to store the k-step motif-based matrices for any reasonable $K$.
To overcome this issue, we avoid \emph{explicitly} constructing the k-step motif-based matrices entirely.
Hence, no additional space is required and we never need to store the actual $k$-step motif-based matrices for $k>1$.
We discuss and show this for any $k$-step motif-based matrix later in this subsection.

Given a k-step motif-based matrix $\mS_t^{(k)}$ for an arbitrary network motif $H_t \in \mathcal{H}$, 
we find an embedding by solving the following optimization problem:
\begin{align} \label{eq:obj-func-local-higher-order-net-rep-motif}
\argmin_{\mU_t^{(k)}\!,\mV_t^{(k)} \!\in \mathcal{C}} \,
\mathbb{D}\,\!
\big(
\,
\mS^{(k)}_t
\,
\| 
\; \Phi \langle\mU_t^{(k)} \! \mV_t^{(k)}
\rangle 
\big),\; \forall k\!=\!1,\!...,\!K
\,\text{and}\,
\;t\!=\!1,\!...,\!T 
\end{align}\noindent
where $\mathbb{D}$ is a generalized Bregman divergence (and quantifies $\approx$ in the HONE embedding model $\mS^{(k)}_t \approx \Phi\langle \mU_t^{(k)}\mV_t^{(k)} \rangle$) with matching linear or non-linear function $\Phi$ and $\mathcal{C}$ is constraints (\eg, non-negativity constraints $\mU \geq 0$, $\mV \geq 0$, orthogonality constraints $\mU^T\mU=\eye$, $\mV^T\mV=\eye$). 
The above optimization problem finds low-rank embedding matrices $\mU_t^{(k)}$ and $\mV_t^{(k)}$ such that $\mS^{(k)}_t \approx \Phi\langle \mU_t^{(k)}\mV_t^{(k)} \rangle$.
The function $\Phi$ allows non-linear relationships between $\mU_t^{(k)}\mV_t^{(k)}$ and $\mS^{(k)}_t$.
Different choices of $\Phi$ and $\mathbb{D}$ yield different HONE embedding models and depend on the distributional assumptions on $\mS^{(k)}_t$.
For instance, minimizing squared loss with an identity link function $\Phi$ yields singular value decomposition corresponding to a Gaussian error model~\cite{golub2012matrix}.
Other choices of $\Phi$ and $\mathbb{D}$ yield other HONE embedding models with different error models such as Poisson, Gamma, or Bernoulli distributions~\cite{collins2002generalization}.

Recall from above that we avoid \emph{explicitly} computing and storing the k-step motif-based matrices from Eq.~\ref{eq:motif-k-step-matrices} as they can densify quickly as $K$ increases and therefore are impractical to store for any large graph and reasonable $K$.
This is accomplished by defining a linear operator corresponding to the $K$-step motif-based matrices that can run in at most $K$ times the linear operator corresponding to the ($1$-step) motif-based matrix.
In particular, many algorithms used to compute low-rank approximations of large sparse matrices~\cite{rokhlin2009randomized,halko2011finding} do not need access to the \emph{explicit matrix}, but only the \emph{linear operator} corresponding to action of the input matrix on vectors. 
For a matrix $\mA$, let $T_{\mA}$ denote the upper bound on the time required to compute $\mA\vx$ for any vector $\vx$. 
We note $T_{\mA} = \mathcal{O}(M)$ where $M = \mathtt{nnz}(\mA)$ always holds and is a useful bound when $\mA$ is sparse. 
Therefore, the time required to compute a rank-$D_{\ell}$ approximation of $\mA$ is $\mathcal{O}(T_{\mA} D_{\ell} \log N  + ND_{\ell}^2 \log N)$ where $N = |V|$.

Now, we can define a linear operator corresponding to the $K$-step motif-based matrices that can run in at most $K$ times the linear operator corresponding to the ($1$-step) motif-based matrix.
We show this for the case of any weighted motif adjacency matrix $\mW$.
Let $T_{\mW}$ be the time required to compute $\mW\vx$, for any vector $\vx$. 
Then, to compute $\mW^K\vx$, we can do the following. 
Let $\vx_0 \leftarrow \vx$ and iteratively compute $\vx_i = \mW\vx_{i-1}$ for $i =1,\ldots,K$. 
This shows that $T_{\mW^K} = \mathcal{O}(K T_{\mW})$.  
This implies that we can compute a rank-$D_{\ell}$ embedding of the $K$-step motif adjacency matrix in time at most 
$\mathcal{O}( K T_{\mW} D_{\ell} \log N  + N D_{\ell}^2 \log N)$ which is at most 
\begin{equation} \label{eq:linear-operator-svd-time-complexity}
\mathcal{O}( KM D_{\ell} \log N  + N D_{\ell}^2 \log N)
\end{equation}\noindent
where $M = \mathtt{nnz}(\mW)$. 
This implies that the time to compute the rank-$D_{\ell}$ embedding grows only linearly with $K$. 
Therefore, no additional space is required and we never need to derive/store the actual k-step motif-based matrices for $k>1$.
Moreover, as shown above, the time complexity grows linearly with $K$ and is therefore efficient.
The time complexity in Eq.~\ref{eq:linear-operator-svd-time-complexity} is for singular value decomposition/eigen-decomposition and hence finds the \emph{best} rank-$D_{\ell}$ approximation~\cite{golub2012matrix}.
However, linear operators can also be defined for other optimization techniques that can be used to compute a rank-$D_{\ell}$ approximation such as stochastic gradient descent, block/cyclic coordinate descent, or alternating least squares.
Thus, the time complexity for computing rank-$D_{\ell}$ embeddings using these optimization techniques will also only increase by a factor of $K$.

Afterwards, the columns of $\mU_t^{(k)}$ are normalized by a function $g : \RR^{N \times N} \!\rightarrow \RR^{N \times N}$ as follows:
\begin{equation} \label{eq:scale-embeddings}
\mU_t^{(k)} \leftarrow g(\mU_t^{(k)}),\quad \text{ for } t=1,\ldots,T \text{ and } k=1,\ldots,K  
\end{equation}
In this work, $g$ is a function that normalizes each column of $\mU_t^{(k)}$ using the Euclidean norm.
The HONE framework is flexible for use with other norms as well and the appropriate norm should be chosen based on the data characteristics and application.

\subsection{Learning \emph{Global} Higher-Order Embeddings} \label{sec:framework-global-Z}
How can we learn a higher-order embedding for an arbitrary graph $G$ that automatically captures the important motifs? 
Obviously, simply concatenating the previous motif embeddings into a single matrix and using this for prediction assumes that each motif is equally important.
However, it is obvious that some motifs are more important than others and the choice of which motifs to use depends on the graph structure and its properties~\cite{prvzulj2007biological,pgd}.
Therefore, instead of assuming all motifs contribute equally to the embedding, we learn a \emph{global} higher-order embedding that automatically captures the important motifs in the embedding without requiring an expert to hand select the most important motifs to use.

For this, we first concatenate the k-step embedding matrices for all $T$ motifs and all $K$ steps:
\begin{equation} \label{eq:concatenate-local-k-step-motif-embeddings} 
\mY \; = \; \Big[ \, \underbrace{\mU_1^{(1)} \;\, \cdots \,\; \mU_T^{(1)}}_{1\text{-step}} 
\;\; \cdots \;\;\, 
\underbrace{\mU_1^{(K)} \;\, \cdots \,\; \mU_T^{(K)}}_{K\text{-steps}} \, \Big]
\end{equation}
\noindent
where $\mY$ is a $N \times TKD_{\ell}$ matrix.
Notice that at this stage, we could simply output $\mY$ as the final motif-based node embeddings and use it for a downstream prediction task such as classification, link prediction, or regression.
However, using $\mY$ directly essentially treats all motifs equally while it is known that some motifs are more important than others and the specific set of important motifs widely depends on the underlying graph structure.
Therefore, by learning node embeddings from $\mY$ we automatically capture the important structure in the data pertaining to certain motifs and avoid having to specify the important motifs for a particular graph by hand.

Given $\mY$ from Eq.~\ref{eq:concatenate-local-k-step-motif-embeddings}, we learn a \emph{global} higher-order network embedding by solving the following:
\begin{align} \label{eq:obj-func-global-higher-order-net-rep-motif}
\argmin_{\mZ,\mH \in \mathcal{C}} \;\,
\mathbb{D} \big(\,\mY\; \Vert \; \Phi \langle\mZ \mH \rangle \big) 
\end{align}\noindent
where $\mZ$ is a $N \times D$ matrix of higher-order node embeddings and $\mH$ is a $D \times TKD_{\ell}$ matrix of the latent $k$-step motif embeddings.
Each row of $\mZ$ is a $D$-dimensional embedding of a node.
Similarly, each column of $\mH$ is an embedding of a latent k-step motif feature (\ie, column of $\mY$) in the same $D$-dimensional space.
In Eq.~\ref{eq:obj-func-global-higher-order-net-rep-motif} we use Frobenius norm which leads to the following minimization problem:
\begin{equation} \label{eq:obj-used-in-exp}
\min_{\mZ, \mH} \; \frac{1}{2} \, \big\|\, \mY - \mZ\mH \big\|_{F}^{2} 
\;=\; \frac{1}{2} \sum_{ij} \big( \mY_{ij} - (\mZ\mH)_{ij}\big)^2
\end{equation}
A similar minimization problem using Frobenius norm is solved for Eq.~\ref{eq:obj-func-local-higher-order-net-rep-motif}.
To solve these minimization problems, we use a fast parallel cyclic coordinate descent-based (CCD) optimization scheme~\cite{hfy12a,pcmf-snam16}. We have also investigated other approaches for solving the above HONE objective function including an autoencoder~\cite{autoencoder,hinton2006reducing} and alternating least squares (ALS)~\cite{zhou2008large} and found similar results.
In addition, it is straightforward to represent the ($k$-step) motif-based matrices 
as a tensor and derive embeddings jointly using Higher Order SVD (Tucker decomposition)~\cite{tucker1966some}, among other higher-order tensor factorization schemes~\cite{kolda2009tensor}.

\subsection{Attribute Diffusion} \label{sec:attribute-diffusion}
Attributes can also be diffused and incorporated into the higher-order node embeddings.
One approach is to use the motif transition probability matrix as follows:
\begin{gather} \label{eq:motif-transition-attribute-diffusion-init}
	\bar{\mX}^{(0)}_t \leftarrow \mX,\;\quad
\mP_t = \mD^{-1}_t \mW_t \nonumber \\
\bar{\mX}^{(k)}_t = \mP_t \bar{\mX}^{(k-1)}_t,\quad \text{ for } k=1,2,\ldots,K \label{eq:motif-transition-attribute-diffusion}
\end{gather}\noindent
where $\mX$ is an $N \times F$ attribute matrix and $\bar{\mX}^{(k)}_t \in \RR^{N \times F}$ is the diffused feature matrix after $k$-steps.
Here $\mP_t$ can be replaced by any of the previous motif-based matrices derived from any motif matrix formulation in Section~\ref{sec:motif-based-matrix-functions}.
More generally, we define \emph{linear attribute diffusion} for HONE as:
\begin{gather} 
	\bar{\mX}^{(0)}_t \leftarrow \mX\; \nonumber \\
\bar{\mX}^{(k)}_t = \Psi \big(\mW_t^{(k)}\big) \bar{\mX}^{(k-1)}_t,\quad \text{ for } k=1,2,\ldots,K \label{eq:motif-attribute-diffusion-general-def}
\end{gather}\noindent
More complex attribute diffusion processes can also be formulated such as the \emph{normalized motif Laplacian attribute diffusion} defined as 
\begin{equation}
\bar{\mX}^{(k)} = (1-\theta)\mL\bar{\mX}^{(k-1)} + \theta\mX,\quad \text{  for } k=1,2,...
\end{equation}
\noindent
where $\mL$ is the normalized motif Laplacian:
\begin{equation}
\mL = \eye - \mD^{\nicefrac{1}{2}}\mW\mD^{\nicefrac{1}{2}}
\end{equation}
The resulting diffused attribute vectors $\bar{\mX} = \big[\;\; \bar{\mX}_1 \;\; \bar{\mX}_2 \;\; \cdots \;\; \big]$ are effectively smoothed by the attributes of related nodes governed by the particular diffusion process.

Afterwards, we incorporate the diffused attribute vectors $\bar{\mX} = \big[\;\; \bar{\mX}_1 \;\; \bar{\mX}_2 \;\; \cdots \;\; \big]$ into the node embeddings given as output in Eq.~\ref{eq:obj-func-global-higher-order-net-rep-motif} by replacing $\mY$ in Eq.~\ref{eq:concatenate-local-k-step-motif-embeddings} with:
\begin{equation}\label{eq:incorporate-diffused-attr-vectors} 
\mY \; = \; \Big[ \, \underbrace{\mU_1^{(1)} \;\, \cdots \,\; \mU_T^{(1)}}_{1\text{-step}} 
\;\; \cdots \;\;\, 
\underbrace{\mU_1^{(K)} \;\, \cdots \,\; \mU_T^{(K)}}_{K\text{-steps}} \,\; \bar{\mX}  \, \Big]
\end{equation}\noindent
Alternatively, we can concatenate $\bar{\mX}$ to $\mZ$, $\big[\, \mZ \,\; \bar{\mX} \, \big]$.
The columns of $\bar{\mX}$ are normalized using Eq.~\ref{eq:scale-embeddings} with the same norm as before.

\subsection{Accumulation Motif Variants} \label{sec:accumulation-motif-variants}
There are also summation-based motif variants.
\begin{equation} \label{eq:sum-motif-adj}
\bar{\mW}^{(K)} = \; \frac{1}{K} \sum_{\ell=1}^{K} \mW^{\ell} \; = \;\, \frac{1}{K}\,\, \Big( \mW + \mW^{2} + \cdots + \mW^{K}\Big)
\end{equation} 
\noindent
where $\bar{\mW}^{(K)}$ is a weighted graph that counts the number of paths of length \emph{up to} $K$.
More interestingly, let 
\begin{equation}\label{eq:sum-motif-transition-matrix}
\bar{\mP}^{(K)} = \; \frac{1}{K} \sum_{\ell=1}^{K} \mP^{\ell} \; = \;\, \frac{1}{K}\,\, \Big( \mP + \mP^{2} + \cdots + \mP^{K}\Big)
\end{equation}
which for instance when $K$=2, indicates the probability of randomly walking from node $i$ to node $j$ in 2 steps.
Alternatively, we can generalize the above as follows:
\begin{align}\label{eq:sum-motif-alpha-decay-generalization-katz}
\bar{\mS}^{(K)} &= \, \frac{1}{K} \sum_{\ell=1}^{K}  \alpha^{\ell} \Psi(\mW^{\ell}) \, \\ \nonumber
&= \,\frac{1}{K} \,\, \Big[\; \alpha \Psi(\mW) \,+\, \alpha^{2} \Psi(\mW^{2})  \,+\, \cdots \,+\, \alpha^{k} \Psi(\mW^{K}) \;\Big]
\end{align}\noindent
where $\alpha$ is a decay factor that penalizes more distant connections.

\section{Analysis} 
\label{sec:theoretical-analysis}
Define $\rho(\mA)$ as the density of $\mA$.
\begin{Claim} \label{eq:claim-motif-adj-matrix-sparsity}
Let $\mW$ denote an arbitrary $k$-vertex motif adjacency matrix where $k>2$, then $\rho(\mA) \geq \rho(\mW)$.
\end{Claim}\noindent
This is straightforward to see as the motif adjacency matrix constructed from the edge frequency of any motif $H$ with more than $k>2$ nodes can be viewed as an additional constraint over the initial adjacency matrix $\mA$. 
Therefore, in the extreme case, if every edge contains at least one occurrence of motif $H$ then $\rho(\mA) = \rho(\mW)$.
However, if there exists at least one edge that does not contain an instance of $H$ then $\rho(\mA) > \rho(\mW)$.
Therefore, $\rho(\mA) \geq \rho(\mW)$.

\subsection{Time Complexity}\label{sec:time-complexity}
Let $M=|E|$, 
$N=|V|$, 
$\Delta=$ the maximum degree, 
$T =$ the number of motifs, 
$K =$ the number of steps, 
$D_{\ell}$ = number of dimensions for each local motif embedding (Section~\ref{sec:framework-local-Y}), and 
$D =$ dimensionality of the final node embeddings (Section~\ref{sec:framework-global-Z}).
\begin{Lemma}
The total time complexity of HONE is
\begin{equation}\label{sec:eq-time-complexity}
\mathcal{O}( M(\Delta_{\rm ub} + KTD_{\ell}) + NDKTD_{\ell})
\end{equation}
\end{Lemma}
\noindent\textsc{Proof}. 
The time complexity of each step is provided below.
For the specific HONE embedding model, we assume $\mathbb{D}$ is squared loss, $\Phi$ is the identity link function, and no hard constraints are imposed on the objective function in Eq.~\ref{eq:obj-func-local-higher-order-net-rep-motif} and Eq.~\ref{eq:obj-func-global-higher-order-net-rep-motif}.

\medskip\noindent\textbf{Weighted motif graphs}: 
To derive the network motif frequencies, we use recent provably accurate estimation methods~\cite{rossi18tnnls,ahmed16bigdata}.
As shown in~\cite{rossi18tnnls,ahmed16bigdata}, we can achieve estimates within a guaranteed level of accuracy and time by setting a few simple parameters in the estimation algorithm.
The time complexity to estimate the frequency of all network motifs up to size 4 is $\mathcal{O}(M\Delta_{\rm ub} )$ in the worst case where $\Delta_{\rm ub}$ is a small constant.
Hence, $\Delta_{\rm ub}$ represents the maximum sampled degree and can be set by the user~\cite{rossi18tnnls,ahmed16bigdata}.

After obtaining the frequencies of the network motifs, we derive a sparse \emph{weighted motif adjacency matrix} for each of the network motifs.
The time complexity for each weighted motif adjacency matrix is at most $\mathcal{O}(M)$ and this is repeated $T$ times for a total time complexity of $\mathcal{O}(MT)$ where $T$ is a small constant.
This gives a total time complexity of $\mathcal{O}(M(T+\Delta_{\rm ub}))$ for this step and thus linear in the number of edges.

\medskip\noindent\textbf{Motif matrix functions}: 
The time complexity of all motif matrix functions $\Psi$ in Section~\ref{sec:motif-based-matrix-functions} is $\mathcal{O}(M)$.
Since $\Psi(\mW_t)$ for $t=1,\ldots,T$, the total time complexity is $\mathcal{O}(MT)$ in the worst case.
By Claim~\ref{eq:claim-motif-adj-matrix-sparsity}, $M \geq M_t$, $\forall t$ where $M_t = \mathtt{nnz}(\mW_t)$ and thus the actual time is likely to be much smaller especially given the rarity of some network motifs in sparse networks such as 4-cliques and 4-cycles.

\medskip\noindent\textbf{Embedding each \textit{k}-step motif graph}: 
For a single weighted motif-based matrix, the time complexity per iteration of cyclic/block coordinate descent~\cite{kim2014algorithms,pcmf-snam16} and stochastic gradient descent~\cite{yun2014nomad,oh2015fast} is at most $\mathcal{O}(MD_{\ell})$ where $D_{\ell} \ll M$.
Recall from Section~\ref{sec:framework-local-Y} that we avoid \emph{explicitly} computing and storing the k-step motif-based matrices by defining a linear operator corresponding to the $K$-step motif-based matrices with a time complexity that is at most $K$ times the linear operator corresponding to the $1$-step motif-based matrix.
Therefore, the total time complexity for learning node embeddings for all $k$-step motif-based matrices is:
\begin{equation} \label{eq:time-complexity-k}
\mathcal{O}\Big( \underbrace{TMD_{\ell} \phantom{\Big(}\!\!\!}_{k=1} \; + \; \underbrace{2(TMD_{\ell})\phantom{\Big(}\!\!\!}_{k=2} 
+ \cdots +
\underbrace{K(TMD_{\ell}) \phantom{\Big(}\!\!\!}_{k=K} \, \Big) \, =\;
\mathcal{O}\big( KTMD_{\ell} \big)
\end{equation}
\noindent

\medskip\noindent\textbf{Global higher-order node embeddings}: 
Afterwards, all $k$-step motif embedding matrices are horizontally concatenated to obtain $\mY$ (Eq.~\ref{eq:concatenate-local-k-step-motif-embeddings}).
Each node embedding matrix is $N \times D_{\ell}$ and there are $K \cdot T$ of them. 
Thus, it takes $\mathcal{O}(NKTD_{\ell})$ time to concatenate them to obtain $\mY$.
Notice that $N \gg KTD_{\ell}$ and therefore this step is linear in the number of nodes $N=|V|$.
Furthermore, the time complexity for normalizing all columns of $\mY$ is $\mathcal{O}(N KTD_{\ell})$ for any normalization function $g$ where each column of $\mY$ is a $N$-dimensional vector.

Given a dense tall-and-skinny matrix $\mY$ of size $N \times KTD_{\ell}$ where $N \gg KTD_{\ell}$, the next step is to learn the higher-order node embedding matrix $\mZ$ and the latent motif embedding matrix $\mH$.
Notice that unlike the higher-order node embeddings above that were derived for each sparse motif-based matrix (for all $K$-steps and $T$ motifs), the matrix $\mY$ is dense with $NKTD_{\ell} = \mathtt{nnz}(\mY)$.
The time complexity per iteration of cyclic/block coordinate descent~\cite{kim2014algorithms,pcmf-snam16} and stochastic gradient descent~\cite{yun2014nomad,oh2015fast} is $\mathcal{O}(DNKTD_{\ell})$ and therefore linear in the number of nodes.

\subsection{Space Complexity} \label{sec:space-complexity}

\begin{Lemma}
The total space complexity of HONE is
\begin{equation}\label{sec:eq-space-complexity}
\mathcal{O}(T(M + NKD_{\ell}) + D(N + TKD_{\ell}))
\end{equation}
\end{Lemma}

\noindent\textsc{Proof}.
The weighted motif adjacency matrices $\mW_1, \ldots, \mW_T$ take at most $\mathcal{O}(MT)$ space.
Similarly, the space complexity of the motif-based matrices derived from any motif matrix function $\Psi$ is at most $\mathcal{O}(MT)$.
Recall that the space required for some motif-based matrices where the motif being encoded is rare will be much less than $\mathcal{O}(MT)$ (Claim~\ref{eq:claim-motif-adj-matrix-sparsity}).
The space complexity of each $k$-step motif embedding is $\mathcal{O}(ND_{\ell})$ and therefore it takes $\mathcal{O}(NTKD_{\ell})$ space for all $k=1,\ldots,K$ and $t=1,\ldots,T$ embedding matrices. 
Storing the higher-order node embedding matrix $\mZ$ takes $\mathcal{O}(ND)$ space and the $k$-step motif embedding matrix $\mH$ is $\mathcal{O}(DTKD_{\ell})$.
Therefore, the total space complexity for $\mZ$ and $\mH$ is 
$\mathcal{O}(ND + DTKD_{\ell}) = \mathcal{O}(D (N + TKD_{\ell}))$.

\section{Experiments} \label{sec:exp}
We investigate five methods from the proposed higher-order network representation learning framework.

\subsection{Experimental Setup} \label{sec:exp-setup}
We compare the proposed HONE variants to five state-of-the-art methods including node2vec \cite{node2vec}, DeepWalk \cite{deepwalk}, LINE \cite{line}, GraRep \cite{grarep}, and Spectral clustering \cite{spectral}.
All methods output $(D=128)$-dimensional node embeddings $\mZ = \big[\, \vz_1 \cdots \vz_N \,\big]^T$ where $\vz_i \in \RR^{D}$.
For LINE, we use 2nd-order-proximity and the number of samples $T=$ 60 million~\cite{line}.
For GraRep, we set $D=128$ and perform a grid search over $K \in \{1,2,3,4\}$~\cite{grarep}.
For DeepWalk, we use 
$R=10$, $L=80$, and $\omega = 10$~\cite{deepwalk}.
For node2vec, we use the same hyperparameters ($D=128$, $R=10$, $L=80$, $\omega = 10$) 
and grid search over $p,q\in \{0.25, 0.50, 1, 2, 4\}$ as mentioned in~\cite{node2vec}. 
For the HONE variants, we set $D=128$ and select the number of steps $K$ automatically via a grid search over $K \in \{1,2,3,4\}$ using $10\%$ of the labeled data.
We use all edge orbits (graphlet automorphisms)~\cite{prvzulj2007biological} that contain 2-4 nodes and set $D_{\ell}=16$ for the local motif embeddings unless otherwise mentioned.
All methods use logistic regression (LR) with an L2 penalty.
The model is selected using 10-fold cross-validation on $10\%$ of the labeled data.
Experiments are repeated for 10 random seed initializations.
All data was obtained from NetworkRepository~\cite{nr} and is publicly available for download at \url{http://networkrepository.com}.

\subsection{Comparison}\label{sec:exp-comparison}
We compare methods from the proposed higher-order network embedding (HONE) framework to other recent embedding methods.
Given a partially observed graph $G$ with a fraction of missing edges, the link prediction task is to predict these missing edges.
We generate a labeled dataset of edges.
Positive examples are obtained by removing $50\%$ of edges randomly, whereas \emph{negative examples} are generated by randomly sampling an equal number of node pairs that are not connected with an edge $(i,j) \not\in E$. 
For each method, we learn embeddings using the remaining graph that consists of only positive examples.
Using the embeddings from each method, we then learn a model to predict whether a given edge in the test set exists in $E$ or not.
To construct edge features from the node embeddings, we use the mean operator defined as $(\vz_i + \vz_j)\big/2$.
For the experiments, we selected networks from a wide range of domains with fundamentally different structural characteristics.
This ensures the key findings observed in this work are more useful/general and apply to networks from a wide variety of domains with different structural characteristics~\cite{network-classification}.

\begin{table}[t!]
\vspace{-2mm}
\centering
\setlength{\tabcolsep}{2.5pt}
\renewcommand{\arraystretch}{1.2} 
\scriptsize
\caption{
AUC results comparing HONE to recent embedding methods across a wide variety of networks from different application domains.
See text for discussion.
}
\vspace{-3mm}
\label{table:link-pred-mean-OP-higher-order-global-layer}
\small
\fontsize{8.0}{9.0}\selectfont
\npdecimalsign{.}
\nprounddigits{3}
\begin{tabularx}{1.0\linewidth}{@{} 
HH
r   
n{1}{3}  
n{1}{3}  
n{1}{3} 
n{1}{3}  
n{1}{3} 
n{1}{3}  
n{1}{3}
c 
H 
@{}
} 

\toprule

&&&   	 
\multicolumn{1}{l}{\rotatebox{80}{\textsf{soc-hamster}}} & 
\multicolumn{1}{l}{\rotatebox{80}{\textsf{rt-twitter-cop}}} &  
\multicolumn{1}{l}{\rotatebox{80}{\textsf{soc-wiki-Vote}}} & 
\multicolumn{1}{l}{\rotatebox{80}{\textsf{tech-routers-rf}}} &  
\multicolumn{1}{l}{\rotatebox{80}{\textsf{facebook-PU}}} & 
\multicolumn{1}{l}{\rotatebox{80}{\textsf{inf-openflights}}} &  
\multicolumn{1}{l}{\rotatebox{80}{\textsf{soc-bitcoinA}}} & 
\multicolumn{1}{l}{\rotatebox{80}{\textsc{Rank}}} & 
\\

\midrule
&&
\textbf{HONE}-$\mW$ (Eq.~\ref{eq:motif-weighted-graph}) &
0.841425  &  
0.842808  & 
0.811242  &  0.862195  &  
0.726349  &  0.910485  &   
0.978651 & 
\textbf{1} & 
\\ 

&& 
\textbf{HONE}-$\mP$ (Eq.~\ref{eq:motif-transition-prob-matrix}) &
0.839928  &   
0.839601  &  
0.811791  &  0.862527  &   
0.724416  &  0.91337  &   
0.980146  &  
\textbf{2} & 
\\ 

&& 
\textbf{HONE}-$\mL$ (Eq.~\ref{eq:simple-motif-laplacian}) &
0.828922  &   
0.841196  &  
0.808381  &  0.857956  &   
0.722109  &  0.905989  &   
0.975094  &  
\textbf{3} & 
\\ 

&& 
\textbf{HONE}-$\hat{\mL}$ (Eq.~\ref{eq:norm-motif-laplacian}) &
0.828654  &   
0.836462  &  
0.802982  &  0.861927  &   
0.72228  &  0.907884  &   
0.975744  & 
\textbf{5} & 
\\ 

&& 
\textbf{HONE}-$\hat{\mL}_{\rm rw}$ (Eq.~\ref{eq:norm-motif-laplacian-rw})
&  	   0.831107  &    
0.833612  &    0.80826  &  0.862827  &   0.722904  &  0.908974  &  0.976269  &  
\textbf{4} & 
\\ 
\midrule

&& 
\textbf{Node2Vec}~\cite{node2vec} & 
0.809905   &   
0.635491   &    
0.72142   &   0.803915   &    
0.701497   &   0.843842   &   
   0.893593 &
\textbf{6} & 
\\

&& \textbf{DeepWalk}~\cite{deepwalk} &  
0.796404   &  
0.621022   &    
0.71043   &   0.79611   &    
0.696045   &   0.836718  & 0.863442 &
\textbf{7} & 
\\

&& \textbf{LINE}~\cite{line} & 
0.752235   &    
0.705533   &    
0.734462   &   0.800182   &    
0.63042   &   0.837354    & 0.779518 &
\textbf{8} & 
\\

&& \textbf{GraRep}~\cite{grarep} &  
0.805437   &  
0.672361   &    
0.743363   &   0.829253   &    
0.702385   &   0.89822     & 0.5594 &
\textbf{9} & 
\\

&& \textbf{Spectral}~\cite{spectral} &  
0.561126   &  
0.699352   &    
0.593448   &   0.602181   &    
0.516098   &   0.605766     & 0.6289 &
\textbf{10} & 
\\

\bottomrule
\end{tabularx}
\npnoround
\vspace{-2mm}
\end{table}

\begin{table}[h!]
\vspace{-1mm}
\centering
\setlength{\tabcolsep}{6.5pt}
\setlength{\tabcolsep}{4.5pt}
\renewcommand{\arraystretch}{1.2} 
\scriptsize
\caption{
Mean gain of the HONE methods over each of the baselines averaged over all graphs.
}
\vspace{-3mm}
\label{table:link-pred-gain}
\small
\fontsize{8.0}{9.0}\selectfont
\npdecimalsign{.}
\nprounddigits{2}
\begin{tabularx}{1.0\linewidth}{@{} 
l
H
ccccc
H 
H
}

\toprule
&&  \multicolumn{1}{l}{\rotatebox{0}{\textbf{Node2Vec}
}} &  
\multicolumn{1}{l}{\rotatebox{0}{\textbf{DeepWalk}
}} &  
\multicolumn{1}{l}{\rotatebox{0}{\textbf{LINE}
}}     &  
\multicolumn{1}{l}{\rotatebox{0}{\textbf{GraRep}
}}   &  
\multicolumn{1}{l}{\rotatebox{0}{\textbf{Spectral}
}} & \\ 
\midrule
\textbf{HONE}-$\mW$            & 
11.02\%  &  12.91\%  &  14.14\%  &  17.52\%  &  42.43\%  &  19.61\%  &  \\
\textbf{HONE}-$\mP$          & 
10.98\%  &  12.86\%  &  14.10\%  &  17.49\%  &  42.39\%  &  19.56\%  &  \\
\textbf{HONE}-$\mL$             & 
10.42\%  &  12.29\%  &  13.51\%  &  16.89\%  &  41.60\%  &  18.94\%  &  \\
\textbf{HONE}-$\hat{\mL}$    & 
10.32\%  &  12.19\%  &  13.42\%  &  16.80\%  &  41.53\%  &  18.85\%  &  \\
\textbf{HONE}-$\hat{\mL}_{\rm rw}$   & 
10.45\%  &  12.33\%  &  13.57\%  &  16.94\%  &  41.74\%  &  19.01\% &  \\

\bottomrule
\end{tabularx}
\npnoround
\vspace{0mm}
\end{table}

The AUC results are provided in Table~\ref{table:link-pred-mean-OP-higher-order-global-layer}.
In all cases, the HONE methods outperform the other embedding methods with an overall mean gain of $19.19\%$ 
(and up to $75.21\%$ gain) across a wide variety of graphs with different characteristics.
Overall, the HONE variants achieve an average gain of $10.64\%$ over node2vec, $12.51\%$ over DeepWalk, $13.75\%$ over LINE, $17.13\%$ over GraRep, and $41.94\%$ over Spectral clustering across all networks.
In all cases, the gain achieved by the proposed HONE variants is significant at $p<0.01$.
We also derive a total ranking of the embedding methods over all graph problems based on mean relative gain.
Results are provided in the last column of Table~\ref{table:link-pred-mean-OP-higher-order-global-layer}. 
Overall, the HONE variants always outperform the five baseline methods across all networks from a wide variety of domains with fundamentally different structural characteristics.
Among the five \textsc{HONE} variants in Table~\ref{table:link-pred-mean-OP-higher-order-global-layer}, we find that \textsc{HONE}-$\mW$ and \textsc{HONE}-$\mP$ perform the best overall.
We also note that GraRep outperforms the other baseline methods on 5 of the 7 graphs
and when soc-bitcoinA is removed GraRep is ranked $6{\rm th}$ outperforming all other baseline methods.
Furthermore, we also provide the mean gain of the HONE methods over each baseline averaged over all graphs in Table~\ref{table:link-pred-gain}.
In other words, an entry in Table~\ref{table:link-pred-gain} represents the mean gain of a HONE method $\mathcal{A}_i$ (row of Table~\ref{table:link-pred-gain}) relative to a baseline method $\mathcal{A}_j$ (column) averaged over all graphs $\mathcal{G}$ used for evaluation.

We also investigated using the concatenated k-step embedding matrix $\mY$ 
directly for link prediction without the additional step described in Section~\ref{sec:framework-global-Z}.
The results were removed for brevity, however, we summarize the findings below.
In most cases, we observed the performance to be better when \emph{global higher-order node embeddings} (Section~\ref{sec:framework-global-Z}) are used as opposed to using the \emph{local node embeddings} from Section~\ref{sec:framework-local-Y} directly for prediction.
We also explored using different optimization schemes for learning the embeddings.
However, we found only minor differences in AUC on most of the graphs investigated.
For instance, on rt-twitter-copen with HONE-P, we found alternating least squares (ALS) and CCD to perform best with $0.865$ AUC followed by $0.864$ using an autoencoder with $f(x) = \nicefrac{1}{1+\exp(-x)}$.

\subsection{Diffusion Variants}
\label{sec:exp-hone-with-attribute-diffusion}
This subsection investigates HONE variants that use attribute diffusion (Section~\ref{sec:attribute-diffusion}).
These methods perform attribute diffusion using the k-step motif matrices (Section~\ref{sec:attribute-diffusion}) and concatenate the resulting diffused features.
Unless otherwise mentioned, we use linear diffusion defined in Eq.~\ref{eq:motif-attribute-diffusion-general-def} with the default hyperparameters (Section~\ref{sec:exp-setup}).
Note the initial matrix $\mX$ described in Section~\ref{sec:attribute-diffusion} represents node motif counts derived by applying relational aggregates (sum, mean, and max) over each nodes local neighborhood and then scaled using Euclidean norm.
We compare the HONE methods \textit{with attribute diffusion} to the HONE methods without diffusion.
Results are reported in Table~\ref{table:link-pred-gain-attr-diffusion}.
The relative gain between each pair of HONE methods is computed for each graph and Table~\ref{table:link-pred-gain-attr-diffusion} reports the mean gain for each pair of HONE methods.
Overall, we observe that HONE with attribute diffusion improves predictive performance in general.
We also investigated other attribute diffusion variants from Section~\ref{sec:attribute-diffusion} and noticed similar results on a few graphs tested.

\begin{table}[h!]
\vspace{-1mm}
\centering
\setlength{\tabcolsep}{6.5pt}
\setlength{\tabcolsep}{1.3pt}
\renewcommand{\arraystretch}{1.2} 
\scriptsize
\caption{
Mean gain of the HONE methods \textit{with attribute diffusion} relative to each of the original HONE methods.
}
\vspace{-3mm}
\label{table:link-pred-gain-attr-diffusion}
\small
\fontsize{8.0}{9.0}\selectfont
\npdecimalsign{.}
\nprounddigits{2}
\begin{tabularx}{1.0\linewidth}{@{} 
l
H
ccccc
H 
H
}

\toprule
&&  \multicolumn{1}{l}{\rotatebox{0}{\textbf{HONE}-$\mW$
}} &  
\multicolumn{1}{l}{\rotatebox{0}{\textbf{HONE}-$\mP$
}} &  
\multicolumn{1}{l}{\rotatebox{0}{ \textbf{HONE}-$\mL$
}}    &  
\multicolumn{1}{l}{\rotatebox{0}{\textbf{HONE}-$\hat{\mL}$
}}&  
\multicolumn{1}{l}{\rotatebox{0}{\textbf{HONE}-$\hat{\mL}_{\rm rw}$
}} & \\ 
\midrule
\textbf{HONE}-$\mW + \bar{\mX}$           & 
0.73\%  &  0.76\%  &  1.30\%  &  1.38\%  &  1.24\%  &  1.08\%  &  \\
\textbf{HONE}-$\mP + \bar{\mX}$          & 
1.54\%  &  1.58\%  &  2.12\%  &  2.20\%  &  2.06\%  &  1.90\%  &  \\
\textbf{HONE}-$\mL + \bar{\mX}$          & 
0.58\%  &  0.62\%  &  1.15\%  &  1.23\%  &  1.09\%  &  0.93\%  &  \\
\textbf{HONE}-$\hat{\mL} + \bar{\mX}$      & 
1.33\%  &  1.37\%  &  1.91\%  &  1.99\%  &  1.85\%  &  1.69\%  &  \\
\textbf{HONE}-$\hat{\mL}_{\rm rw} + \bar{\mX}$ & 
1.23\%  &  1.27\%  &  1.81\%  &  1.88\%  &  1.74\%  &  1.58\%  &  \\

\bottomrule
\end{tabularx}
\npnoround
\vspace{-3mm}
\end{table}

\subsection{Runtime \& Scalability} 
\label{sec:exp-runtime-scalability}
\begin{figure}[h!]
\centering
\includegraphics[width=0.60\linewidth]{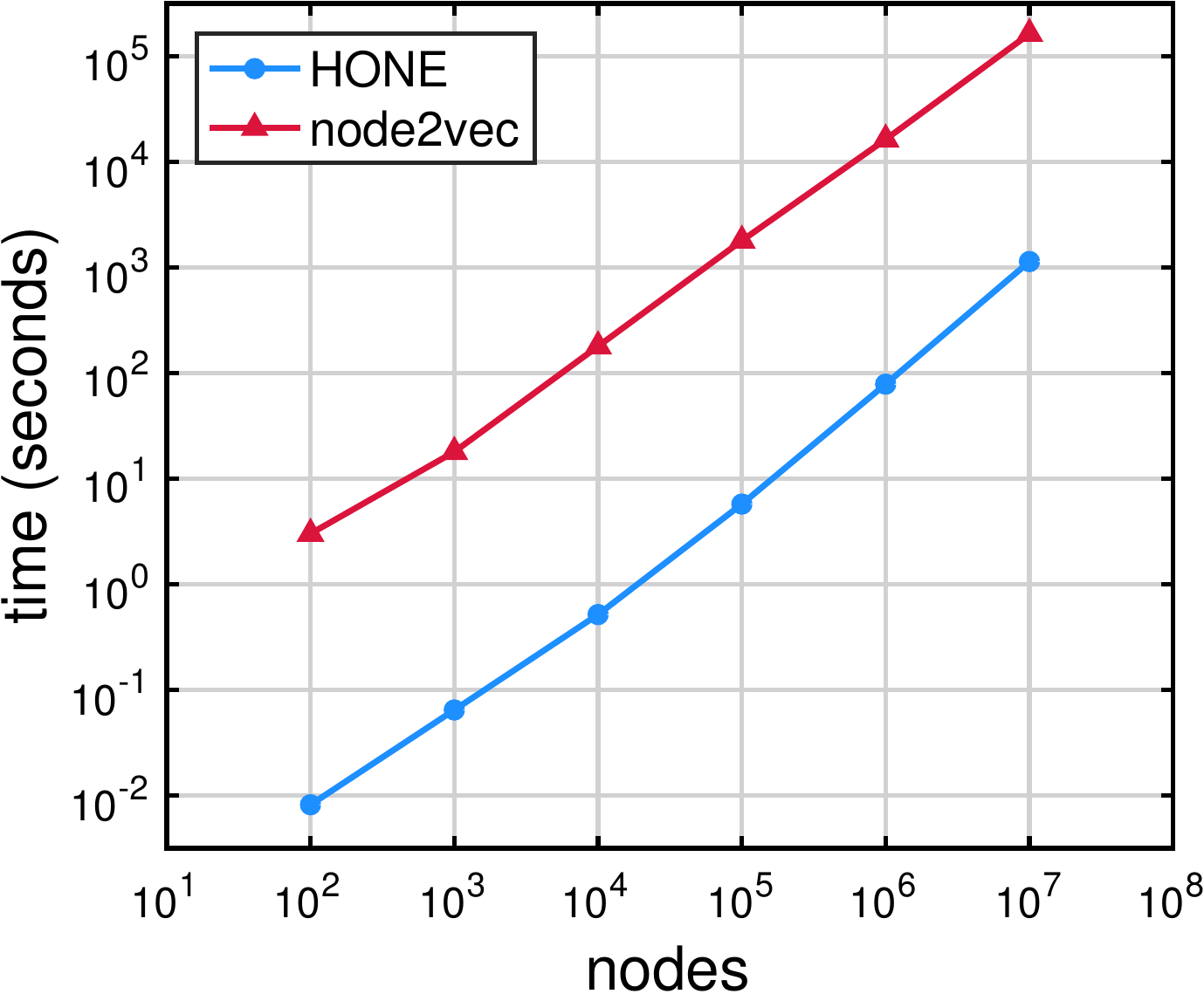}
\caption{
Runtime comparison on  Erd\protect\"{o}s-R\protect\'{e}nyi graphs with an average degree of 10.
HONE is shown to be scalable and orders of magnitude faster than node2vec.
See text for discussion.
}
\label{fig:exp-runtime-ER}
\end{figure}

\noindent
To evaluate the runtime performance and scalability of the proposed framework, 
we learn node embeddings for Erd\"{o}s-R\'{e}nyi graphs of increasing size (from 100 to 10 million nodes) such that each graph has an average degree of 10.
In Figure~\ref{fig:exp-runtime-ER}, we observe that HONE is fast and scales linearly as the number of nodes increases.
In addition, we also compare the runtime performance of HONE against node2vec~\cite{node2vec} since it performed best among the baselines (Table~\ref{table:link-pred-mean-OP-higher-order-global-layer}).
For the HONE variant, we use HONE-P with $K=2$. 
Default parameters are used for each method.
In Figure~\ref{fig:exp-runtime-ER}, HONE is shown to be 
significantly faster and more scalable than node2vec as the number of nodes increases.
In particular, node2vec takes 1.8 days (45.3 hours) for 10 million nodes,
while HONE finishes in only 19 minutes as shown in Figure~\ref{fig:exp-runtime-ER}.
Strikingly, this is $143$ times faster than node2vec.

\newcommand{\subsec}[1]{\smallskip\noindent\textbf{#1}:}

\section{Related Work} \label{sec:related-work}\noindent
Related research is categorized below.

\subsec{Higher-order network analysis}
This paper introduces the problem of \emph{higher-order network embedding} and proposes a general computational framework for learning such higher-order node embeddings.
There has been one recent approach that used network motifs as base features for network representation learning~\cite{deepGL}.
However, that approach is fundamentally different from the proposed framework as it focuses on learning inductive relational functions that represent compositions of relational operators applied to a base feature. 
Other methods use high-order network properties (such as graphlet frequencies) as features for graph classification~\cite{vishwanathan2010graph},  community detection~\cite{arenas2008motif,benson2016higher},
and visualization and exploratory analysis~\cite{pgd}. 
However, this work focuses on network representation learning using network motifs (\eg, orbit frequencies).
In particular, the goal of this work is to learn higher-order node embeddings from the graph for use in a downstream prediction task.

\subsec{Node embeddings}
There has been a lot of interest recently in learning node (and edge~\cite{ahmed2017roles}) embeddings
from large-scale networks automatically~\cite{node2vec,deepwalk,line,deepGL,role2vec,ComE,struc2vec}.
See~\cite{rossi12jair} for an early survey on graph representation learning.
Recent node embedding methods~\cite{deepwalk,node2vec,line,struc2vec,ComE} have largely been based on the popular skip-gram model~\cite{skipgram-old,cheng2006n} originally introduced for learning vector representations of words in text.
These methods all use random walks to gather a sequence of node ids which are then used to learn node embeddings~\cite{deepwalk,node2vec,line,struc2vec,ComE}.
In particular, DeepWalk~\cite{deepwalk} applied the successful word embedding framework called word2vec~\cite{skipgram-old} to embed the nodes such that the co-occurrence frequencies of pairs in short random walks are preserved. 
Recently, node2vec~\cite{node2vec} adapted DeepWalk~\cite{deepwalk} by introducing hyperparameters that tune the depth and breadth of the random walks.
GraRep~\cite{grarep} is a generalization of LINE~\cite{line} that incorporates node neighborhood information beyond 2-hops.
These approaches are becoming increasingly popular and have been shown to outperform a number of existing methods.
Graph Convolutional Networks (GCNs) adapt CNNs to graphs using the simple Laplacian and spectral convolutions with a form of aggregation over the neighbors~\cite{henaff2015deep,defferrard2016convolutional,kipf2016ssl,CNN-graphs}.
These node embedding methods may also benefit from ideas developed in this work including the weighted motif Laplacian matrices described in Section~\ref{sec:motif-based-matrix-functions}.
Other work has focused on incremental methods for spectral clustering~\cite{chen2015incremental}.
Similar techniques can be used to derive incremental methods for updating HONE; however, this is outside the scope of this paper.

There is also another related body of work focused on attributed graphs.
Recently, Huang~\etal~\cite{huang2017label} proposed a label informed embedding method for attributed networks.
This approach assumes the graph is labeled and uses this information to improve predictive performance.
However, this paper does not focus on attributed-based embeddings and therefore is significantly different.
First and foremost, while the class of HONE models are able to support attributed graphs via attributed diffusion, this work does not focus on such graphs.
Moreover, HONE does not require attributes or class labels on the nodes.

Heterogeneous networks~\cite{shi2014hetesim} have also been recently considered~\cite{chang2015heterogeneous,dong2017metapath2vec} as well as attributed networks with labels~\cite{huang2017label,huang2017accelerated}.
Huang~\etal~\shortcite{huang2017label} proposed an approach for attributed networks with labels whereas Yang~\etal~\shortcite{yang2015network} used text features to learn node representations.
Liang~\etal~\shortcite{liang2017seano} proposed a semi-supervised approach for networks with outliers. 
Bojchevski~\etal~\shortcite{bojchevski2017deep} proposed an unsupervised rank-based approach.
There has also been some recent work on 
semi-supervised network embeddings~\cite{Planetoid,kipf2016ssl} and methods for improving the learned representations~\cite{weston2008deep,scarselli2009graph,wang2016structural}.
A few work have begun to explore the problem of learning node embeddings from temporal networks~\cite{rossi2013dbmm-wsdm,
saha2018models,rahman2018dylink2vec}.
All of these approaches \emph{approximate} the dynamic network as a sequence of discrete static snapshot graphs.
More recently, methods have been proposed that use temporal random walks to avoid the information loss of previous discrete approximation methods~\cite{CTDNE-WWW18}.
This work is different from the problem discussed in this paper.

\subsec{Role-based embeddings}
Many recent node embedding methods have attempted to capture roles~\cite{roles2015-tkde} by preserving the notion of structural equivalence~\cite{everett1985role} or the relaxed notion of structural similarity~\cite{roles2015-tkde}.
Examples of node roles include nodes acting as hubs, bridges (acting as gate-keepers), near-cliques, and star-edges.
Over the previous decade, there have been many role-based embedding (role discovery) methods that automatically learn node embeddings from graphs; see~\cite{roles2015-tkde} for a survey.
These methods are some of the earliest such embedding (representation learning) methods for graphs.
More recently, most approaches have been based on traditional random walks and thus are unable to capture roles (structural equivalence or structural similarity) and instead capture the notion of communities~\cite{deepwalk,node2vec,struc2vec,ComE}.
In particular, these methods embed nodes in a similar way that are close to one another in the graph and therefore are largely capturing the notion of communities as opposed to roles.
Instead, nodes that are structurally similar (\ie, share similar general connectivity/subgraph patterns) should be embedded in a similar way, independent of their proximity to one another in the graph.
Recently, an approach called role2vec was proposed that learns role-based node embeddings by first mapping each node to a type via a function and then uses the notion of attributed (typed) random walks to derive role-based embeddings for the nodes that capture structural similarity~\cite{role2vec}.
This approach was shown to generalize many existing random walk-based methods.

\subsec{Graph embeddings}
Methods such as DeepWalk and node2vec learn embeddings for nodes in a graph.
On the other hand, there have recently been methods that learn embeddings for entire graphs \cite{Duvenaud15,lee17-Deep-Graph-Attention}.
These methods can be used for graph-level tasks like graph classification. 
In particular, methods such as Random Walk Kernel~\cite{vishwanathan2010graph}, Deep Graph Kernel~\cite{Yanardag15}, and SkipGraph \cite{skipgraph} make use of random walks to learn embeddings for entire graphs.
More recently, a \emph{graph attention model} was proposed in~\cite{lee17-Deep-Graph-Attention} and used for graph classification.
Other work has also focused on developing graph embeding methods for attributed molecular graphs~\cite{coley2017convolutional}.

\subsec{Improving autocorrelation}
This work is also related to recent methods for improving autocorrelation and classification performance by creating new links~\cite{gallagher2008using} and even relevance in search engines~\cite{lassez2008ranking}.
Intuitively, HONE naturally estimates weights between new previously unobserved edges based on k-step motif patterns.
In particular, new links are explicitly created between nodes in the k-step motif matrices.

\section{Conclusion} \label{sec:conc}
In this work, we proposed \emph{Higher-Order Network Embeddings} (HONE), a new class of embedding methods that use network motifs to learn embeddings 
based on higher-order connectivity patterns.
We describe a general computational framework for learning such higher-order network embeddings that is flexible with many interchangeable components.
The experimental results demonstrated the effectiveness of learning higher-order network representations as HONE achieves a mean relative gain in AUC of $19\%$ across all other methods and networks from a wide variety of application domains.
Future work will investigate the framework using other useful motif-based matrix formulations.

\balance
\bibliographystyle{ACM-Reference-Format}
\bibliography{paper}  


\begin{thebibliography}{68}


\ifx \showCODEN    \undefined \def \showCODEN     #1{\unskip}     \fi
\ifx \showDOI      \undefined \def \showDOI       #1{#1}\fi
\ifx \showISBNx    \undefined \def \showISBNx     #1{\unskip}     \fi
\ifx \showISBNxiii \undefined \def \showISBNxiii  #1{\unskip}     \fi
\ifx \showISSN     \undefined \def \showISSN      #1{\unskip}     \fi
\ifx \showLCCN     \undefined \def \showLCCN      #1{\unskip}     \fi
\ifx \shownote     \undefined \def \shownote      #1{#1}          \fi
\ifx \showarticletitle \undefined \def \showarticletitle #1{#1}   \fi
\ifx \showURL      \undefined \def \showURL       {\relax}        \fi
\providecommand\bibfield[2]{#2}
\providecommand\bibinfo[2]{#2}
\providecommand\natexlab[1]{#1}
\providecommand\showeprint[2][]{arXiv:#2}

\bibitem[\protect\citeauthoryear{Ahmed, Neville, Rossi, and Duffield}{Ahmed
  et~al\mbox{.}}{2015}]%
        {pgd}
\bibfield{author}{\bibinfo{person}{Nesreen~K. Ahmed}, \bibinfo{person}{Jennifer
  Neville}, \bibinfo{person}{Ryan~A. Rossi}, {and} \bibinfo{person}{Nick
  Duffield}.} \bibinfo{year}{2015}\natexlab{}.
\newblock \showarticletitle{Efficient Graphlet Counting for Large Networks}. In
  \bibinfo{booktitle}{\emph{ICDM}}. \bibinfo{pages}{10}.
\newblock


\bibitem[\protect\citeauthoryear{Ahmed, Rossi, Willke, and Zhou}{Ahmed
  et~al\mbox{.}}{2017a}]%
        {ahmed2017roles}
\bibfield{author}{\bibinfo{person}{Nesreen~K. Ahmed}, \bibinfo{person}{Ryan~A.
  Rossi}, \bibinfo{person}{Theodore~L. Willke}, {and} \bibinfo{person}{Rong
  Zhou}.} \bibinfo{year}{2017}\natexlab{a}.
\newblock \showarticletitle{{Edge Role Discovery via Higher-Order Structures}}.
  In \bibinfo{booktitle}{\emph{PAKDD}}. \bibinfo{pages}{291--303}.
\newblock


\bibitem[\protect\citeauthoryear{Ahmed, Rossi, Zhou, Lee, Kong, Willke, and
  Eldardiry}{Ahmed et~al\mbox{.}}{2017b}]%
        {ahmed17Gen-Deep-Graph-Learning}
\bibfield{author}{\bibinfo{person}{Nesreen~K. Ahmed}, \bibinfo{person}{Ryan~A.
  Rossi}, \bibinfo{person}{Rong Zhou}, \bibinfo{person}{John~Boaz Lee},
  \bibinfo{person}{Xiangnan Kong}, \bibinfo{person}{Theodore~L. Willke}, {and}
  \bibinfo{person}{Hoda Eldardiry}.} \bibinfo{year}{2017}\natexlab{b}.
\newblock \showarticletitle{A Framework for Generalizing Graph-based
  Representation Learning Methods}. In
  \bibinfo{booktitle}{\emph{arXiv:1709.04596}}. \bibinfo{pages}{8}.
\newblock


\bibitem[\protect\citeauthoryear{Ahmed, Rossi, Zhou, Lee, Kong, Willke, and
  Eldardiry}{Ahmed et~al\mbox{.}}{2018}]%
        {role2vec}
\bibfield{author}{\bibinfo{person}{Nesreen~K. Ahmed}, \bibinfo{person}{Ryan~A.
  Rossi}, \bibinfo{person}{Rong Zhou}, \bibinfo{person}{John~Boaz Lee},
  \bibinfo{person}{Xiangnan Kong}, \bibinfo{person}{Theodore~L. Willke}, {and}
  \bibinfo{person}{Hoda Eldardiry}.} \bibinfo{year}{2018}\natexlab{}.
\newblock \showarticletitle{Learning Role-based Graph Embeddings}. In
  \bibinfo{booktitle}{\emph{arXiv:1802.02896}}.
\newblock


\bibitem[\protect\citeauthoryear{Ahmed, Willke, and Rossi}{Ahmed
  et~al\mbox{.}}{2016}]%
        {ahmed16bigdata}
\bibfield{author}{\bibinfo{person}{Nesreen~K. Ahmed},
  \bibinfo{person}{Theodore~L. Willke}, {and} \bibinfo{person}{Ryan~A. Rossi}.}
  \bibinfo{year}{2016}\natexlab{}.
\newblock \showarticletitle{Estimation of Local Subgraph Counts}. In
  \bibinfo{booktitle}{\emph{BigData}}. \bibinfo{pages}{586--595}.
\newblock


\bibitem[\protect\citeauthoryear{Arenas, Fernandez, Fortunato, and
  Gomez}{Arenas et~al\mbox{.}}{2008}]%
        {arenas2008motif}
\bibfield{author}{\bibinfo{person}{Alex Arenas}, \bibinfo{person}{Alberto
  Fernandez}, \bibinfo{person}{Santo Fortunato}, {and} \bibinfo{person}{Sergio
  Gomez}.} \bibinfo{year}{2008}\natexlab{}.
\newblock \showarticletitle{Motif-based communities in complex networks}.
\newblock \bibinfo{journal}{\emph{Journal of Physics A: Mathematical and
  Theoretical}} \bibinfo{volume}{41}, \bibinfo{number}{22}
  (\bibinfo{year}{2008}), \bibinfo{pages}{224001}.
\newblock


\bibitem[\protect\citeauthoryear{Benson, Gleich, and Leskovec}{Benson
  et~al\mbox{.}}{2016}]%
        {benson2016higher}
\bibfield{author}{\bibinfo{person}{Austin~R Benson}, \bibinfo{person}{David~F
  Gleich}, {and} \bibinfo{person}{Jure Leskovec}.}
  \bibinfo{year}{2016}\natexlab{}.
\newblock \showarticletitle{Higher-order organization of complex networks}.
\newblock \bibinfo{journal}{\emph{Science}} \bibinfo{volume}{353},
  \bibinfo{number}{6295} (\bibinfo{year}{2016}), \bibinfo{pages}{163--166}.
\newblock


\bibitem[\protect\citeauthoryear{Bojchevski and G{\"u}nnemann}{Bojchevski and
  G{\"u}nnemann}{2017}]%
        {bojchevski2017deep}
\bibfield{author}{\bibinfo{person}{Aleksandar Bojchevski} {and}
  \bibinfo{person}{Stephan G{\"u}nnemann}.} \bibinfo{year}{2017}\natexlab{}.
\newblock \showarticletitle{Deep Gaussian Embedding of Attributed Graphs:
  Unsupervised Inductive Learning via Ranking}.
\newblock \bibinfo{journal}{\emph{arXiv:1707.03815}} (\bibinfo{year}{2017}).
\newblock


\bibitem[\protect\citeauthoryear{Canning, Ingram, Nowak-Wolff, Ortiz, Ahmed,
  Rossi, Schmitt, and Soundarajan}{Canning et~al\mbox{.}}{2018}]%
        {network-classification}
\bibfield{author}{\bibinfo{person}{James~P. Canning}, \bibinfo{person}{Emma~E.
  Ingram}, \bibinfo{person}{Sammantha Nowak-Wolff}, \bibinfo{person}{Adriana~M.
  Ortiz}, \bibinfo{person}{Nesreen~K. Ahmed}, \bibinfo{person}{Ryan~A. Rossi},
  \bibinfo{person}{Karl R.~B. Schmitt}, {and} \bibinfo{person}{Sucheta
  Soundarajan}.} \bibinfo{year}{2018}\natexlab{}.
\newblock \showarticletitle{Network Classification and Categorization}. In
  \bibinfo{booktitle}{\emph{International Conference on Complex Networks
  (CompleNet)}}.
\newblock


\bibitem[\protect\citeauthoryear{Cao, Lu, and Xu}{Cao et~al\mbox{.}}{2015}]%
        {grarep}
\bibfield{author}{\bibinfo{person}{Shaosheng Cao}, \bibinfo{person}{Wei Lu},
  {and} \bibinfo{person}{Qiongkai Xu}.} \bibinfo{year}{2015}\natexlab{}.
\newblock \showarticletitle{GraRep: Learning graph representations with global
  structural information}. In \bibinfo{booktitle}{\emph{CIKM}}. ACM,
  \bibinfo{pages}{891--900}.
\newblock


\bibitem[\protect\citeauthoryear{Cavallari, Zheng, Cai, Chang, and
  Cambria}{Cavallari et~al\mbox{.}}{2017}]%
        {ComE}
\bibfield{author}{\bibinfo{person}{Sandro Cavallari},
  \bibinfo{person}{Vincent~W Zheng}, \bibinfo{person}{Hongyun Cai},
  \bibinfo{person}{Kevin Chen-Chuan Chang}, {and} \bibinfo{person}{Erik
  Cambria}.} \bibinfo{year}{2017}\natexlab{}.
\newblock \showarticletitle{Learning community embedding with community
  detection and node embedding on graphs}. In \bibinfo{booktitle}{\emph{CIKM}}.
  \bibinfo{pages}{377--386}.
\newblock


\bibitem[\protect\citeauthoryear{Chang, Han, Tang, Qi, Aggarwal, and
  Huang}{Chang et~al\mbox{.}}{2015}]%
        {chang2015heterogeneous}
\bibfield{author}{\bibinfo{person}{Shiyu Chang}, \bibinfo{person}{Wei Han},
  \bibinfo{person}{Jiliang Tang}, \bibinfo{person}{Guo-Jun Qi},
  \bibinfo{person}{Charu~C Aggarwal}, {and} \bibinfo{person}{Thomas~S Huang}.}
  \bibinfo{year}{2015}\natexlab{}.
\newblock \showarticletitle{Heterogeneous network embedding via deep
  architectures}. In \bibinfo{booktitle}{\emph{SIGKDD}}.
  \bibinfo{pages}{119--128}.
\newblock


\bibitem[\protect\citeauthoryear{Chen, Zhang, Hasan, and Hero}{Chen
  et~al\mbox{.}}{2015}]%
        {chen2015incremental}
\bibfield{author}{\bibinfo{person}{Pin-Yu Chen}, \bibinfo{person}{Baichuan
  Zhang}, \bibinfo{person}{Mohammad~Al Hasan}, {and} \bibinfo{person}{Alfred~O
  Hero}.} \bibinfo{year}{2015}\natexlab{}.
\newblock \showarticletitle{Incremental method for spectral clustering of
  increasing orders}. In \bibinfo{booktitle}{\emph{arXiv:1512.07349}}.
\newblock


\bibitem[\protect\citeauthoryear{Cheng, Greaves, and Warren}{Cheng
  et~al\mbox{.}}{2006}]%
        {cheng2006n}
\bibfield{author}{\bibinfo{person}{Winnie Cheng}, \bibinfo{person}{Chris
  Greaves}, {and} \bibinfo{person}{Martin Warren}.}
  \bibinfo{year}{2006}\natexlab{}.
\newblock \showarticletitle{From n-gram to skipgram to concgram}.
\newblock \bibinfo{journal}{\emph{Int. J. of Corp. Linguistics}}
  \bibinfo{volume}{11}, \bibinfo{number}{4} (\bibinfo{year}{2006}),
  \bibinfo{pages}{411--433}.
\newblock


\bibitem[\protect\citeauthoryear{Coley, Barzilay, Green, Jaakkola, and
  Jensen}{Coley et~al\mbox{.}}{2017}]%
        {coley2017convolutional}
\bibfield{author}{\bibinfo{person}{Connor~W Coley}, \bibinfo{person}{Regina
  Barzilay}, \bibinfo{person}{William~H Green}, \bibinfo{person}{Tommi~S
  Jaakkola}, {and} \bibinfo{person}{Klavs~F Jensen}.}
  \bibinfo{year}{2017}\natexlab{}.
\newblock \showarticletitle{Convolutional Embedding of Attributed Molecular
  Graphs for Physical Property Prediction}.
\newblock \bibinfo{journal}{\emph{J. Chem. Info. \& Mod.}}
  (\bibinfo{year}{2017}).
\newblock


\bibitem[\protect\citeauthoryear{Collins, Dasgupta, and Schapire}{Collins
  et~al\mbox{.}}{2002}]%
        {collins2002generalization}
\bibfield{author}{\bibinfo{person}{Michael Collins}, \bibinfo{person}{Sanjoy
  Dasgupta}, {and} \bibinfo{person}{Robert~E Schapire}.}
  \bibinfo{year}{2002}\natexlab{}.
\newblock \showarticletitle{A generalization of principal components analysis
  to the exponential family}. In \bibinfo{booktitle}{\emph{NIPS}}.
  \bibinfo{pages}{617--624}.
\newblock


\bibitem[\protect\citeauthoryear{Defferrard, Bresson, and
  Vandergheynst}{Defferrard et~al\mbox{.}}{2016}]%
        {defferrard2016convolutional}
\bibfield{author}{\bibinfo{person}{Micha{\"e}l Defferrard},
  \bibinfo{person}{Xavier Bresson}, {and} \bibinfo{person}{Pierre
  Vandergheynst}.} \bibinfo{year}{2016}\natexlab{}.
\newblock \showarticletitle{Convolutional neural networks on graphs with fast
  localized spectral filtering}. In \bibinfo{booktitle}{\emph{Advances in
  Neural Information Processing Systems}}. \bibinfo{pages}{3844--3852}.
\newblock


\bibitem[\protect\citeauthoryear{Dong, Chawla, and Swami}{Dong
  et~al\mbox{.}}{2017}]%
        {dong2017metapath2vec}
\bibfield{author}{\bibinfo{person}{Yuxiao Dong}, \bibinfo{person}{Nitesh~V
  Chawla}, {and} \bibinfo{person}{Ananthram Swami}.}
  \bibinfo{year}{2017}\natexlab{}.
\newblock \showarticletitle{metapath2vec: Scalable Representation Learning for
  Heterogeneous Networks}. In \bibinfo{booktitle}{\emph{SIGKDD}}.
\newblock


\bibitem[\protect\citeauthoryear{Duvenaud, Maclaurin, Aguilera-Iparraguirre,
  Bombarell, Hirzel, Aspuru-Guzik, and Adams}{Duvenaud et~al\mbox{.}}{2015}]%
        {Duvenaud15}
\bibfield{author}{\bibinfo{person}{David~K. Duvenaud}, \bibinfo{person}{Dougal
  Maclaurin}, \bibinfo{person}{Jorge Aguilera-Iparraguirre},
  \bibinfo{person}{Rafael Bombarell}, \bibinfo{person}{Timothy Hirzel},
  \bibinfo{person}{Alan Aspuru-Guzik}, {and} \bibinfo{person}{Ryan~P. Adams}.}
  \bibinfo{year}{2015}\natexlab{}.
\newblock \showarticletitle{Convolutional networks on graphs for learning
  molecular fingerprints}. In \bibinfo{booktitle}{\emph{NIPS}}.
\newblock


\bibitem[\protect\citeauthoryear{Everett}{Everett}{1985}]%
        {everett1985role}
\bibfield{author}{\bibinfo{person}{M.G. Everett}.}
  \bibinfo{year}{1985}\natexlab{}.
\newblock \showarticletitle{Role similarity and complexity in social networks}.
\newblock \bibinfo{journal}{\emph{Social Networks}} \bibinfo{volume}{7},
  \bibinfo{number}{4} (\bibinfo{year}{1985}), \bibinfo{pages}{353--359}.
\newblock


\bibitem[\protect\citeauthoryear{Fortunato}{Fortunato}{2010}]%
        {fortunato2010community}
\bibfield{author}{\bibinfo{person}{S. Fortunato}.}
  \bibinfo{year}{2010}\natexlab{}.
\newblock \showarticletitle{Community detection in graphs}.
\newblock \bibinfo{journal}{\emph{Phy. Rep.}} \bibinfo{volume}{486},
  \bibinfo{number}{3-5} (\bibinfo{year}{2010}).
\newblock


\bibitem[\protect\citeauthoryear{Gallagher, Tong, Eliassi-Rad, and
  Faloutsos}{Gallagher et~al\mbox{.}}{2008}]%
        {gallagher2008using}
\bibfield{author}{\bibinfo{person}{Brian Gallagher}, \bibinfo{person}{Hanghang
  Tong}, \bibinfo{person}{Tina Eliassi-Rad}, {and} \bibinfo{person}{Christos
  Faloutsos}.} \bibinfo{year}{2008}\natexlab{}.
\newblock \showarticletitle{Using ghost edges for classification in sparsely
  labeled networks}. In \bibinfo{booktitle}{\emph{SIGKDD}}.
\newblock


\bibitem[\protect\citeauthoryear{Golub and Van~Loan}{Golub and
  Van~Loan}{2012}]%
        {golub2012matrix}
\bibfield{author}{\bibinfo{person}{Gene~H Golub} {and}
  \bibinfo{person}{Charles~F Van~Loan}.} \bibinfo{year}{2012}\natexlab{}.
\newblock \bibinfo{booktitle}{\emph{Matrix computations}}.
\newblock \bibinfo{publisher}{JHU Press}.
\newblock


\bibitem[\protect\citeauthoryear{Grover and Leskovec}{Grover and
  Leskovec}{2016}]%
        {node2vec}
\bibfield{author}{\bibinfo{person}{Aditya Grover} {and} \bibinfo{person}{Jure
  Leskovec}.} \bibinfo{year}{2016}\natexlab{}.
\newblock \showarticletitle{node2vec: Scalable feature learning for networks}.
  In \bibinfo{booktitle}{\emph{SIGKDD}}. \bibinfo{pages}{855--864}.
\newblock


\bibitem[\protect\citeauthoryear{Halko, Martinsson, and Tropp}{Halko
  et~al\mbox{.}}{2011}]%
        {halko2011finding}
\bibfield{author}{\bibinfo{person}{Nathan Halko}, \bibinfo{person}{Per-Gunnar
  Martinsson}, {and} \bibinfo{person}{Joel~A Tropp}.}
  \bibinfo{year}{2011}\natexlab{}.
\newblock \showarticletitle{Finding structure with randomness: Probabilistic
  algorithms for constructing approximate matrix decompositions}.
\newblock \bibinfo{journal}{\emph{SIAM review}} \bibinfo{volume}{53},
  \bibinfo{number}{2} (\bibinfo{year}{2011}), \bibinfo{pages}{217--288}.
\newblock


\bibitem[\protect\citeauthoryear{Henaff, Bruna, and LeCun}{Henaff
  et~al\mbox{.}}{2015}]%
        {henaff2015deep}
\bibfield{author}{\bibinfo{person}{Mikael Henaff}, \bibinfo{person}{Joan
  Bruna}, {and} \bibinfo{person}{Yann LeCun}.} \bibinfo{year}{2015}\natexlab{}.
\newblock \showarticletitle{Deep convolutional networks on graph-structured
  data}.
\newblock \bibinfo{journal}{\emph{arXiv:1506.05163}} (\bibinfo{year}{2015}).
\newblock


\bibitem[\protect\citeauthoryear{Hinton and Salakhutdinov}{Hinton and
  Salakhutdinov}{2006}]%
        {hinton2006reducing}
\bibfield{author}{\bibinfo{person}{Geoffrey~E Hinton} {and}
  \bibinfo{person}{Ruslan~R Salakhutdinov}.} \bibinfo{year}{2006}\natexlab{}.
\newblock \showarticletitle{Reducing the dimensionality of data with neural
  networks}.
\newblock \bibinfo{journal}{\emph{science}} \bibinfo{volume}{313},
  \bibinfo{number}{5786} (\bibinfo{year}{2006}), \bibinfo{pages}{504--507}.
\newblock


\bibitem[\protect\citeauthoryear{Huang, Li, and Hu}{Huang
  et~al\mbox{.}}{2017a}]%
        {huang2017accelerated}
\bibfield{author}{\bibinfo{person}{Xiao Huang}, \bibinfo{person}{Jundong Li},
  {and} \bibinfo{person}{Xia Hu}.} \bibinfo{year}{2017}\natexlab{a}.
\newblock \showarticletitle{Accelerated attributed network embedding}. In
  \bibinfo{booktitle}{\emph{SDM}}.
\newblock


\bibitem[\protect\citeauthoryear{Huang, Li, and Hu}{Huang
  et~al\mbox{.}}{2017b}]%
        {huang2017label}
\bibfield{author}{\bibinfo{person}{Xiao Huang}, \bibinfo{person}{Jundong Li},
  {and} \bibinfo{person}{Xia Hu}.} \bibinfo{year}{2017}\natexlab{b}.
\newblock \showarticletitle{Label informed attributed network embedding}. In
  \bibinfo{booktitle}{\emph{WSDM}}.
\newblock


\bibitem[\protect\citeauthoryear{Kim, He, and Park}{Kim et~al\mbox{.}}{2014}]%
        {kim2014algorithms}
\bibfield{author}{\bibinfo{person}{Jingu Kim}, \bibinfo{person}{Yunlong He},
  {and} \bibinfo{person}{Haesun Park}.} \bibinfo{year}{2014}\natexlab{}.
\newblock \showarticletitle{Algorithms for nonnegative matrix and tensor
  factorizations: A unified view based on block coordinate descent framework}.
\newblock \bibinfo{journal}{\emph{Journal of Global Optimization}}
  \bibinfo{volume}{58}, \bibinfo{number}{2} (\bibinfo{year}{2014}),
  \bibinfo{pages}{285--319}.
\newblock


\bibitem[\protect\citeauthoryear{Kipf and Welling}{Kipf and Welling}{2017}]%
        {kipf2016ssl}
\bibfield{author}{\bibinfo{person}{Thomas~N Kipf} {and} \bibinfo{person}{Max
  Welling}.} \bibinfo{year}{2017}\natexlab{}.
\newblock \showarticletitle{Semi-supervised classification with graph
  convolutional networks}. In \bibinfo{booktitle}{\emph{ICLR}}.
\newblock


\bibitem[\protect\citeauthoryear{Kolda and Bader}{Kolda and Bader}{2009}]%
        {kolda2009tensor}
\bibfield{author}{\bibinfo{person}{Tamara~G Kolda} {and}
  \bibinfo{person}{Brett~W Bader}.} \bibinfo{year}{2009}\natexlab{}.
\newblock \showarticletitle{Tensor decompositions and applications}.
\newblock \bibinfo{journal}{\emph{SIAM review}} \bibinfo{volume}{51},
  \bibinfo{number}{3} (\bibinfo{year}{2009}), \bibinfo{pages}{455--500}.
\newblock


\bibitem[\protect\citeauthoryear{Lassez, Rossi, and Jeev}{Lassez
  et~al\mbox{.}}{2008}]%
        {lassez2008ranking}
\bibfield{author}{\bibinfo{person}{Jean-Louis Lassez}, \bibinfo{person}{Ryan
  Rossi}, {and} \bibinfo{person}{Kumar Jeev}.} \bibinfo{year}{2008}\natexlab{}.
\newblock \showarticletitle{Ranking Links on the Web: Search and Surf Engines}.
\newblock \bibinfo{journal}{\emph{IEA/AIE}} (\bibinfo{year}{2008}),
  \bibinfo{pages}{199--208}.
\newblock


\bibitem[\protect\citeauthoryear{Lee and Kong}{Lee and Kong}{2017}]%
        {skipgraph}
\bibfield{author}{\bibinfo{person}{John~Boaz Lee} {and} \bibinfo{person}{X.
  Kong}.} \bibinfo{year}{2017}\natexlab{}.
\newblock \showarticletitle{{Skip-Graph: Learning graph embeddings with an
  encoder-decoder model}}. In \bibinfo{booktitle}{\emph{ICLR OpenReview}}.
\newblock


\bibitem[\protect\citeauthoryear{Lee, Rossi, and Kong}{Lee
  et~al\mbox{.}}{2018}]%
        {lee17-Deep-Graph-Attention}
\bibfield{author}{\bibinfo{person}{John~Boaz Lee}, \bibinfo{person}{Ryan
  Rossi}, {and} \bibinfo{person}{Xiangnan Kong}.}
  \bibinfo{year}{2018}\natexlab{}.
\newblock \showarticletitle{Graph Classification using Structural Attention}.
  In \bibinfo{booktitle}{\emph{SIGKDD}}.
\newblock


\bibitem[\protect\citeauthoryear{Liang, Jacobs, and Parthasarathy}{Liang
  et~al\mbox{.}}{2017}]%
        {liang2017seano}
\bibfield{author}{\bibinfo{person}{Jiongqian Liang}, \bibinfo{person}{Peter
  Jacobs}, {and} \bibinfo{person}{Srinivasan Parthasarathy}.}
  \bibinfo{year}{2017}\natexlab{}.
\newblock \showarticletitle{SEANO: Semi-supervised Embedding in Attributed
  Networks with Outliers}. In \bibinfo{booktitle}{\emph{arXiv:1703.08100}}.
\newblock


\bibitem[\protect\citeauthoryear{Mikolov, Chen, Corrado, and Dean}{Mikolov
  et~al\mbox{.}}{2013}]%
        {skipgram-old}
\bibfield{author}{\bibinfo{person}{Tomas Mikolov}, \bibinfo{person}{Kai Chen},
  \bibinfo{person}{Greg Corrado}, {and} \bibinfo{person}{Jeffrey Dean}.}
  \bibinfo{year}{2013}\natexlab{}.
\newblock \showarticletitle{Efficient estimation of word representations in
  vector space}. In \bibinfo{booktitle}{\emph{ICLR Workshop}}.
  \bibinfo{pages}{10}.
\newblock


\bibitem[\protect\citeauthoryear{Nguyen, Lee, Rossi, Ahmed, Koh, and
  Kim}{Nguyen et~al\mbox{.}}{2018}]%
        {CTDNE-WWW18}
\bibfield{author}{\bibinfo{person}{Giang~Hoang Nguyen},
  \bibinfo{person}{John~Boaz Lee}, \bibinfo{person}{Ryan~A. Rossi},
  \bibinfo{person}{Nesreen~K. Ahmed}, \bibinfo{person}{Eunyee Koh}, {and}
  \bibinfo{person}{Sungchul Kim}.} \bibinfo{year}{2018}\natexlab{}.
\newblock \showarticletitle{Continuous-Time Dynamic Network Embeddings}. In
  \bibinfo{booktitle}{\emph{WWW BigNet}}.
\newblock


\bibitem[\protect\citeauthoryear{Niepert, Ahmed, and Kutzkov}{Niepert
  et~al\mbox{.}}{2016}]%
        {CNN-graphs}
\bibfield{author}{\bibinfo{person}{Mathias Niepert}, \bibinfo{person}{Mohamed
  Ahmed}, {and} \bibinfo{person}{Konstantin Kutzkov}.}
  \bibinfo{year}{2016}\natexlab{}.
\newblock \showarticletitle{Learning Convolutional Neural Networks for Graphs}.
  In \bibinfo{booktitle}{\emph{arXiv:1605.05273}}.
\newblock


\bibitem[\protect\citeauthoryear{Oh, Han, Yu, and Jiang}{Oh
  et~al\mbox{.}}{2015}]%
        {oh2015fast}
\bibfield{author}{\bibinfo{person}{Jinoh Oh}, \bibinfo{person}{Wook-Shin Han},
  \bibinfo{person}{Hwanjo Yu}, {and} \bibinfo{person}{Xiaoqian Jiang}.}
  \bibinfo{year}{2015}\natexlab{}.
\newblock \showarticletitle{Fast and robust parallel SGD matrix factorization}.
  In \bibinfo{booktitle}{\emph{SIGKDD}}. ACM, \bibinfo{pages}{865--874}.
\newblock


\bibitem[\protect\citeauthoryear{Perozzi, Al-Rfou, and Skiena}{Perozzi
  et~al\mbox{.}}{2014}]%
        {deepwalk}
\bibfield{author}{\bibinfo{person}{Bryan Perozzi}, \bibinfo{person}{Rami
  Al-Rfou}, {and} \bibinfo{person}{Steven Skiena}.}
  \bibinfo{year}{2014}\natexlab{}.
\newblock \showarticletitle{Deepwalk: Online learning of social
  representations}. In \bibinfo{booktitle}{\emph{SIGKDD}}.
  \bibinfo{pages}{701--710}.
\newblock


\bibitem[\protect\citeauthoryear{Pr{\v{z}}ulj}{Pr{\v{z}}ulj}{2007}]%
        {prvzulj2007biological}
\bibfield{author}{\bibinfo{person}{Nata{\v{s}}a Pr{\v{z}}ulj}.}
  \bibinfo{year}{2007}\natexlab{}.
\newblock \showarticletitle{Biological network comparison using graphlet degree
  distribution}.
\newblock \bibinfo{journal}{\emph{Bioinfo.}} \bibinfo{volume}{23},
  \bibinfo{number}{2} (\bibinfo{year}{2007}), \bibinfo{pages}{e177--e183}.
\newblock


\bibitem[\protect\citeauthoryear{Rahman, Saha, Hasan, Xu, and Reddy}{Rahman
  et~al\mbox{.}}{2018}]%
        {rahman2018dylink2vec}
\bibfield{author}{\bibinfo{person}{Mahmudur Rahman},
  \bibinfo{person}{Tanay~Kumar Saha}, \bibinfo{person}{Mohammad~Al Hasan},
  \bibinfo{person}{Kevin~S Xu}, {and} \bibinfo{person}{Chandan~K Reddy}.}
  \bibinfo{year}{2018}\natexlab{}.
\newblock \showarticletitle{DyLink2Vec: Effective Feature Representation for
  Link Prediction in Dynamic Networks}.
\newblock \bibinfo{journal}{\emph{arXiv:1804.05755}} (\bibinfo{year}{2018}).
\newblock


\bibitem[\protect\citeauthoryear{Ribeiro, Saverese, and Figueiredo}{Ribeiro
  et~al\mbox{.}}{2017}]%
        {struc2vec}
\bibfield{author}{\bibinfo{person}{Leonardo~F.R. Ribeiro},
  \bibinfo{person}{Pedro~H.P. Saverese}, {and} \bibinfo{person}{Daniel~R.
  Figueiredo}.} \bibinfo{year}{2017}\natexlab{}.
\newblock \showarticletitle{Struc2Vec: Learning Node Representations from
  Structural Identity}. In \bibinfo{booktitle}{\emph{SIGKDD}}.
\newblock


\bibitem[\protect\citeauthoryear{Rokhlin, Szlam, and Tygert}{Rokhlin
  et~al\mbox{.}}{2009}]%
        {rokhlin2009randomized}
\bibfield{author}{\bibinfo{person}{Vladimir Rokhlin}, \bibinfo{person}{Arthur
  Szlam}, {and} \bibinfo{person}{Mark Tygert}.}
  \bibinfo{year}{2009}\natexlab{}.
\newblock \showarticletitle{A randomized algorithm for principal component
  analysis}.
\newblock \bibinfo{journal}{\emph{SIAM J. Matrix Anal. Appl.}}
  \bibinfo{volume}{31}, \bibinfo{number}{3} (\bibinfo{year}{2009}).
\newblock


\bibitem[\protect\citeauthoryear{Rossi and Ahmed}{Rossi and Ahmed}{2015a}]%
        {nr}
\bibfield{author}{\bibinfo{person}{Ryan~A. Rossi} {and}
  \bibinfo{person}{Nesreen~K. Ahmed}.} \bibinfo{year}{2015}\natexlab{a}.
\newblock \showarticletitle{The Network Data Repository with Interactive Graph
  Analytics and Visualization}. In \bibinfo{booktitle}{\emph{AAAI}}.
  \bibinfo{pages}{4292--4293}.
\newblock
\urldef\tempurl%
\url{http://networkrepository.com}
\showURL{%
\tempurl}


\bibitem[\protect\citeauthoryear{Rossi and Ahmed}{Rossi and Ahmed}{2015b}]%
        {roles2015-tkde}
\bibfield{author}{\bibinfo{person}{Ryan~A. Rossi} {and}
  \bibinfo{person}{Nesreen~K. Ahmed}.} \bibinfo{year}{2015}\natexlab{b}.
\newblock \showarticletitle{Role Discovery in Networks}.
\newblock \bibinfo{journal}{\emph{Transactions on Knowledge and Data
  Engineering}} \bibinfo{volume}{27}, \bibinfo{number}{4}
  (\bibinfo{date}{April} \bibinfo{year}{2015}), \bibinfo{pages}{1112--1131}.
\newblock


\bibitem[\protect\citeauthoryear{Rossi, Gallagher, Neville, and
  Henderson}{Rossi et~al\mbox{.}}{2013}]%
        {rossi2013dbmm-wsdm}
\bibfield{author}{\bibinfo{person}{Ryan~A. Rossi}, \bibinfo{person}{Brian
  Gallagher}, \bibinfo{person}{Jennifer Neville}, {and} \bibinfo{person}{Keith
  Henderson}.} \bibinfo{year}{2013}\natexlab{}.
\newblock \showarticletitle{Modeling Dynamic Behavior in Large Evolving
  Graphs}. In \bibinfo{booktitle}{\emph{Proceedings of the Sixth ACM
  International Conference on Web Search and Data Mining}}.
  \bibinfo{pages}{667--676}.
\newblock


\bibitem[\protect\citeauthoryear{Rossi, McDowell, Aha, and Neville}{Rossi
  et~al\mbox{.}}{2012}]%
        {rossi12jair}
\bibfield{author}{\bibinfo{person}{Ryan~A. Rossi}, \bibinfo{person}{Luke~K.
  McDowell}, \bibinfo{person}{David~W. Aha}, {and} \bibinfo{person}{Jennifer
  Neville}.} \bibinfo{year}{2012}\natexlab{}.
\newblock \showarticletitle{Transforming graph data for statistical relational
  learning}.
\newblock \bibinfo{journal}{\emph{Journal of Artificial Intelligence Research}}
  \bibinfo{volume}{45}, \bibinfo{number}{1} (\bibinfo{year}{2012}),
  \bibinfo{pages}{363--441}.
\newblock


\bibitem[\protect\citeauthoryear{Rossi and Zhou}{Rossi and Zhou}{2016}]%
        {pcmf-snam16}
\bibfield{author}{\bibinfo{person}{Ryan~A. Rossi} {and} \bibinfo{person}{Rong
  Zhou}.} \bibinfo{year}{2016}\natexlab{}.
\newblock \showarticletitle{{Parallel Collective Factorization for Modeling
  Large Heterogeneous Networks}}. In \bibinfo{booktitle}{\emph{Social Network
  Analysis and Mining}}. \bibinfo{pages}{30}.
\newblock


\bibitem[\protect\citeauthoryear{Rossi, Zhou, and Ahmed}{Rossi
  et~al\mbox{.}}{2018a}]%
        {deepGL}
\bibfield{author}{\bibinfo{person}{Ryan~A. Rossi}, \bibinfo{person}{Rong Zhou},
  {and} \bibinfo{person}{Nesreen~K. Ahmed}.} \bibinfo{year}{2018}\natexlab{a}.
\newblock \showarticletitle{Deep Inductive Network Representation Learning}. In
  \bibinfo{booktitle}{\emph{Proceedings of the 3rd International Workshop on
  Learning Representations for Big Networks (WWW BigNet)}}.
\newblock


\bibitem[\protect\citeauthoryear{Rossi, Zhou, and Ahmed}{Rossi
  et~al\mbox{.}}{2018b}]%
        {rossi18tnnls}
\bibfield{author}{\bibinfo{person}{Ryan~A. Rossi}, \bibinfo{person}{Rong Zhou},
  {and} \bibinfo{person}{Nesreen~K. Ahmed}.} \bibinfo{year}{2018}\natexlab{b}.
\newblock \showarticletitle{Estimation of Graphlet Counts in Massive Networks}.
  In \bibinfo{booktitle}{\emph{IEEE Transactions on Neural Networks and
  Learning Systems (TNNLS)}}. \bibinfo{pages}{1--14}.
\newblock


\bibitem[\protect\citeauthoryear{Rumelhart, Hinton, and Williams}{Rumelhart
  et~al\mbox{.}}{1986}]%
        {autoencoder}
\bibfield{author}{\bibinfo{person}{David~E Rumelhart},
  \bibinfo{person}{Geoffrey~E Hinton}, {and} \bibinfo{person}{Ronald~J
  Williams}.} \bibinfo{year}{1986}\natexlab{}.
\newblock \showarticletitle{Learning internal representations by
  backpropagating errors}.
\newblock \bibinfo{journal}{\emph{Nature}}  \bibinfo{volume}{323}
  (\bibinfo{year}{1986}), \bibinfo{pages}{533--536}.
\newblock


\bibitem[\protect\citeauthoryear{Saha, Williams, Hasan, Joty, and Varberg}{Saha
  et~al\mbox{.}}{2018}]%
        {saha2018models}
\bibfield{author}{\bibinfo{person}{Tanay~Kumar Saha}, \bibinfo{person}{Thomas
  Williams}, \bibinfo{person}{Mohammad~Al Hasan}, \bibinfo{person}{Shafiq
  Joty}, {and} \bibinfo{person}{Nicholas~K Varberg}.}
  \bibinfo{year}{2018}\natexlab{}.
\newblock \showarticletitle{Models for Capturing Temporal Smoothness in
  Evolving Networks for Learning Latent Representation of Nodes}. In
  \bibinfo{booktitle}{\emph{arXiv:1804.05816}}.
\newblock


\bibitem[\protect\citeauthoryear{Scarselli, Gori, Tsoi, Hagenbuchner, and
  Monfardini}{Scarselli et~al\mbox{.}}{2009}]%
        {scarselli2009graph}
\bibfield{author}{\bibinfo{person}{Franco Scarselli}, \bibinfo{person}{Marco
  Gori}, \bibinfo{person}{Ah~Chung Tsoi}, \bibinfo{person}{Markus
  Hagenbuchner}, {and} \bibinfo{person}{Gabriele Monfardini}.}
  \bibinfo{year}{2009}\natexlab{}.
\newblock \showarticletitle{The graph neural network model}.
\newblock \bibinfo{journal}{\emph{IEEE Transactions on Neural Networks}}
  \bibinfo{volume}{20}, \bibinfo{number}{1} (\bibinfo{year}{2009}),
  \bibinfo{pages}{61--80}.
\newblock


\bibitem[\protect\citeauthoryear{Shi, Kong, Huang, Philip, and Wu}{Shi
  et~al\mbox{.}}{2014}]%
        {shi2014hetesim}
\bibfield{author}{\bibinfo{person}{Chuan Shi}, \bibinfo{person}{Xiangnan Kong},
  \bibinfo{person}{Yue Huang}, \bibinfo{person}{S~Yu Philip}, {and}
  \bibinfo{person}{Bin Wu}.} \bibinfo{year}{2014}\natexlab{}.
\newblock \showarticletitle{{HeteSim: A General Framework for Relevance Measure
  in Heterogeneous Networks}}.
\newblock \bibinfo{journal}{\emph{TKDE}} \bibinfo{volume}{26},
  \bibinfo{number}{10} (\bibinfo{year}{2014}), \bibinfo{pages}{2479--2492}.
\newblock


\bibitem[\protect\citeauthoryear{Tang, Qu, Wang, Zhang, Yan, and Mei}{Tang
  et~al\mbox{.}}{2015}]%
        {line}
\bibfield{author}{\bibinfo{person}{Jian Tang}, \bibinfo{person}{Meng Qu},
  \bibinfo{person}{Mingzhe Wang}, \bibinfo{person}{Ming Zhang},
  \bibinfo{person}{Jun Yan}, {and} \bibinfo{person}{Qiaozhu Mei}.}
  \bibinfo{year}{2015}\natexlab{}.
\newblock \showarticletitle{{LINE: Large-scale Information Network Embedding}}.
  In \bibinfo{booktitle}{\emph{WWW}}. \bibinfo{pages}{1067--1077}.
\newblock


\bibitem[\protect\citeauthoryear{Tang and Liu}{Tang and Liu}{2011}]%
        {spectral}
\bibfield{author}{\bibinfo{person}{Lei Tang} {and} \bibinfo{person}{Huan Liu}.}
  \bibinfo{year}{2011}\natexlab{}.
\newblock \showarticletitle{Leveraging social media networks for
  classification}.
\newblock \bibinfo{journal}{\emph{Data Mining and Knowledge Discovery}}
  \bibinfo{volume}{23}, \bibinfo{number}{3} (\bibinfo{year}{2011}),
  \bibinfo{pages}{447--478}.
\newblock


\bibitem[\protect\citeauthoryear{Tucker}{Tucker}{1966}]%
        {tucker1966some}
\bibfield{author}{\bibinfo{person}{Ledyard~R Tucker}.}
  \bibinfo{year}{1966}\natexlab{}.
\newblock \showarticletitle{Some mathematical notes on three-mode factor
  analysis}.
\newblock \bibinfo{journal}{\emph{Psychometrika}} \bibinfo{volume}{31},
  \bibinfo{number}{3} (\bibinfo{year}{1966}), \bibinfo{pages}{279--311}.
\newblock


\bibitem[\protect\citeauthoryear{Vishwanathan, Schraudolph, Kondor, and
  Borgwardt}{Vishwanathan et~al\mbox{.}}{2010}]%
        {vishwanathan2010graph}
\bibfield{author}{\bibinfo{person}{S~Vichy~N Vishwanathan},
  \bibinfo{person}{Nicol~N Schraudolph}, \bibinfo{person}{Risi Kondor}, {and}
  \bibinfo{person}{Karsten~M Borgwardt}.} \bibinfo{year}{2010}\natexlab{}.
\newblock \showarticletitle{Graph kernels}.
\newblock \bibinfo{journal}{\emph{JMLR}}  \bibinfo{volume}{11}
  (\bibinfo{year}{2010}), \bibinfo{pages}{1201--1242}.
\newblock


\bibitem[\protect\citeauthoryear{Wang, Cui, and Zhu}{Wang
  et~al\mbox{.}}{2016}]%
        {wang2016structural}
\bibfield{author}{\bibinfo{person}{Daixin Wang}, \bibinfo{person}{Peng Cui},
  {and} \bibinfo{person}{Wenwu Zhu}.} \bibinfo{year}{2016}\natexlab{}.
\newblock \showarticletitle{Structural deep network embedding}. In
  \bibinfo{booktitle}{\emph{SIGKDD}}. \bibinfo{pages}{1225--1234}.
\newblock


\bibitem[\protect\citeauthoryear{Weston, Ratle, and Collobert}{Weston
  et~al\mbox{.}}{2008}]%
        {weston2008deep}
\bibfield{author}{\bibinfo{person}{Jason Weston},
  \bibinfo{person}{Fr{\'e}d{\'e}ric Ratle}, {and} \bibinfo{person}{Ronan
  Collobert}.} \bibinfo{year}{2008}\natexlab{}.
\newblock \showarticletitle{Deep learning via semi-supervised embedding}. In
  \bibinfo{booktitle}{\emph{ICML}}. \bibinfo{pages}{1168--1175}.
\newblock


\bibitem[\protect\citeauthoryear{Yanardag and Vishwanathan}{Yanardag and
  Vishwanathan}{2015}]%
        {Yanardag15}
\bibfield{author}{\bibinfo{person}{Pinar Yanardag} {and}
  \bibinfo{person}{S.~V.~N. Vishwanathan}.} \bibinfo{year}{2015}\natexlab{}.
\newblock \showarticletitle{Deep Graph Kernels}. In
  \bibinfo{booktitle}{\emph{SIGKDD}}.
\newblock


\bibitem[\protect\citeauthoryear{Yang, Liu, Zhao, Sun, and Chang}{Yang
  et~al\mbox{.}}{2015}]%
        {yang2015network}
\bibfield{author}{\bibinfo{person}{Cheng Yang}, \bibinfo{person}{Zhiyuan Liu},
  \bibinfo{person}{Deli Zhao}, \bibinfo{person}{Maosong Sun}, {and}
  \bibinfo{person}{Edward~Y Chang}.} \bibinfo{year}{2015}\natexlab{}.
\newblock \showarticletitle{Network Representation Learning with Rich Text
  Information.}. In \bibinfo{booktitle}{\emph{IJCAI}}.
\newblock


\bibitem[\protect\citeauthoryear{Yang, Cohen, and Salakhutdinov}{Yang
  et~al\mbox{.}}{2016}]%
        {Planetoid}
\bibfield{author}{\bibinfo{person}{Zhilin Yang}, \bibinfo{person}{William~W
  Cohen}, {and} \bibinfo{person}{Ruslan Salakhutdinov}.}
  \bibinfo{year}{2016}\natexlab{}.
\newblock \showarticletitle{Revisiting semi-supervised learning with graph
  embeddings}.
\newblock \bibinfo{journal}{\emph{arXiv:1603.08861}} (\bibinfo{year}{2016}).
\newblock


\bibitem[\protect\citeauthoryear{Yu, Hsieh, Si, and Dhillon}{Yu
  et~al\mbox{.}}{2012}]%
        {hfy12a}
\bibfield{author}{\bibinfo{person}{Hsiang-Fu Yu}, \bibinfo{person}{Cho-Jui
  Hsieh}, \bibinfo{person}{Si Si}, {and} \bibinfo{person}{Inderjit~S.
  Dhillon}.} \bibinfo{year}{2012}\natexlab{}.
\newblock \showarticletitle{Scalable Coordinate Descent Approaches to Parallel
  Matrix Factorization for Recommender Systems}. In
  \bibinfo{booktitle}{\emph{IEEE International Conference of Data Mining}}.
\newblock


\bibitem[\protect\citeauthoryear{Yun, Yu, Hsieh, Vishwanathan, and Dhillon}{Yun
  et~al\mbox{.}}{2014}]%
        {yun2014nomad}
\bibfield{author}{\bibinfo{person}{Hyokun Yun}, \bibinfo{person}{Hsiang-Fu Yu},
  \bibinfo{person}{Cho-Jui Hsieh}, \bibinfo{person}{SVN Vishwanathan}, {and}
  \bibinfo{person}{Inderjit Dhillon}.} \bibinfo{year}{2014}\natexlab{}.
\newblock \showarticletitle{NOMAD: Non-locking, stOchastic Multi-machine
  algorithm for Asynchronous and Decentralized matrix completion}.
\newblock \bibinfo{journal}{\emph{VLDB}} \bibinfo{volume}{7},
  \bibinfo{number}{11} (\bibinfo{year}{2014}), \bibinfo{pages}{975--986}.
\newblock


\bibitem[\protect\citeauthoryear{Zhou, Wilkinson, Schreiber, and Pan}{Zhou
  et~al\mbox{.}}{2008}]%
        {zhou2008large}
\bibfield{author}{\bibinfo{person}{Yunhong Zhou}, \bibinfo{person}{Dennis
  Wilkinson}, \bibinfo{person}{Robert Schreiber}, {and} \bibinfo{person}{Rong
  Pan}.} \bibinfo{year}{2008}\natexlab{}.
\newblock \showarticletitle{Large-scale parallel collaborative filtering for
  the netflix prize}.
\newblock \bibinfo{journal}{\emph{LNCS}}  \bibinfo{volume}{5034}
  (\bibinfo{year}{2008}).
\newblock


\end{thebibliography}

\end{document}